%% file: tacl_paper.tex
\documentclass[11pt,a4paper]{article}
\usepackage{times,latexsym}
\usepackage{url}

\usepackage[T1]{fontenc}
\usepackage{graphicx}

\usepackage[hang,flushmargin]{footmisc}
\usepackage{subcaption}
\usepackage{multirow}
\usepackage{array}
\usepackage{booktabs}
\usepackage{pgfplots}
\pgfplotsset{compat=1.17}
\usetikzlibrary{positioning}
\usepackage{pdflscape}

\usepackage[acceptedWithA]{tacl2021v1}

\usepackage{xspace,mfirstuc,tabulary}

\newif\iftaclinstructions
\taclinstructionsfalse
\iftaclinstructions
\renewcommand{\confidential}{}
\renewcommand{\anonsubtext}{(No author info supplied here, for consistency with
TACL-submission anonymization requirements)}
\newcommand{\instr}
\fi

\iftaclpubformat

\else

\fi
\input{preamble}

\title{Model Directions, Not Words:\\ Mechanistic Topic Models Using Sparse Autoencoders}

\author{%
  \bf
  Carolina Zheng\textsuperscript{\normalfont *\ensuremath{\diamond}}\quad Nicolas Beltran‑Velez\textsuperscript{\normalfont *\ensuremath{\diamond}}\quad Sweta Karlekar\textsuperscript{\normalfont *\ensuremath{\diamond}}\quad Claudia Shi\textsuperscript{\normalfont\ensuremath{\diamond}}\\[0.5ex]
  \bf Achille Nazaret\textsuperscript{\normalfont\ensuremath{\diamond}}\quad Asif Mallik\textsuperscript{\normalfont\ensuremath{\ddagger}}\quad Amir Feder\textsuperscript{\normalfont\ensuremath{\diamond\dagger}}\quad David M. Blei\textsuperscript{\normalfont\ensuremath{\diamond}}\\[1ex]
  \rm \textsuperscript{\ensuremath{\diamond}}Columbia University, USA\quad \textsuperscript{\ensuremath{\dagger}}Google Research, USA\quad \textsuperscript{\ensuremath{\ddagger}}Independent\\
  \rm {\fontfamily{pcr}\selectfont\{cz2539, nb2838, ssk2275\}@columbia.edu}%
}
\begin{document}
\maketitle

\begingroup
  \renewcommand\thefootnote{\fnsymbol{footnote}}%
  \setlength{\footnotemargin}{0pt}%
  \setcounter{footnote}{1}%
  \footnotetext{\hspace{0.12em}Equal contribution.}%
\endgroup
\setcounter{footnote}{0}%

\begin{abstract}

Traditional topic models are effective at uncovering latent themes in large text collections. However, due to their reliance on bag-of-words representations, they struggle to capture
semantically abstract features. 
While some neural variants use richer representations, they are similarly constrained by expressing topics as word lists, which limits their ability to articulate complex topics.
We introduce Mechanistic Topic Models (MTMs), a class of topic models that operate on interpretable features learned by sparse autoencoders (SAEs). By defining topics over this semantically rich space, MTMs can reveal deeper conceptual themes with expressive feature descriptions. Moreover, uniquely among topic models, MTMs enable controllable text generation using topic steering vectors.
To properly evaluate MTM topics against word list approaches, we propose \textit{topic judge}, an LLM-based pairwise comparison evaluation framework.
Across eight datasets, MTMs match or exceed traditional and neural baselines on coherence metrics, are consistently preferred by topic judge, and enable effective LLM steering.\footnote{\hspace{0.12em}\scriptsize{\renewcommand{\ttdefault}{pcr}\href{https://github.com/blei-lab/mechanistic-topic-models}{\texttt{github.com/blei-lab/mechanistic-topic-models}}}}
\end{abstract}

\input{sections/1-introduction}
\input{sections/2-related_work}
\input{sections/3-background}
\input{sections/4-method}
\input{sections/5-empirical_studies}
\input{sections/6-discussion}

\clearpage
\flushbottom

\section*{Acknowledgments}
We thank the reviewers and action editor for their thoughtful feedback, which has substantially improved the paper.
David M. Blei was supported by NSF IIS-2127869, NSF DMS-2311108, ONR N000142412243, and the Simons Foundation.

\bibliography{tacl2021}
\bibliographystyle{acl_natbib}

\clearpage
\appendix
\captionsetup{font=appendixten,labelfont=appendixten,textfont=appendixten,skip=3pt}
\captionsetup[subfigure]{font=appendixten,labelfont=appendixten,textfont=appendixten,skip=3pt}
\setlength{\textfloatsep}{8pt plus 2pt minus 2pt}
\setlength{\floatsep}{8pt plus 2pt minus 2pt}
\setlength{\intextsep}{8pt plus 2pt minus 2pt}
\setlength{\dbltextfloatsep}{8pt plus 2pt minus 2pt}
\setlength{\dblfloatsep}{8pt plus 2pt minus 2pt}
\input{sections/appendices/mtm_implementation_details}
\input{sections/appendices/datasets}
\input{sections/appendices/baseline_implementations}
\input{sections/appendices/bayesian_optimization}
\input{sections/appendices/steering_experiment_details}
\input{sections/appendices/all_prompts}
\input{sections/appendices/topic_judge}
\input{sections/appendices/human_evaluation}
\input{sections/appendices/topic_refinement_example}
\input{sections/appendices/topic-alignment-details}

\input{sections/appendices/additional_results}

\input{sections/appendices/topic_examples}
\end{document}

%% file: preamble.tex
\usepackage[utf8]{inputenc}%

\usepackage{colortbl}

\usepackage{listings}
\usepackage[scaled=0.85]{inconsolata}
\lstdefinestyle{rawoutput}{
  basicstyle=\ttfamily\scriptsize,
  breaklines=true,
  columns=fullflexible,
  frame=single,
  rulecolor=\color{gray!50},
  backgroundcolor=\color{gray!10},
  showstringspaces=false
}
\lstdefinestyle{poem}{
  basicstyle=\ttfamily\scriptsize,
  breaklines=true,
  frame=none,
  showstringspaces=false
}

\usepackage{array, tabularx}
\newcolumntype{C}{>{\centering\arraybackslash}X}
\newcolumntype{Y}{>{\centering\arraybackslash}X}
\usepackage{siunitx}
\usepackage{multirow}
\usepackage{lipsum}           %

\usepackage{stmaryrd}

\newcommand{\extrasmall}{\fontsize{7.2pt}{8.2pt}\selectfont}

\definecolor{cb0base}{HTML}{B39DEB}%
\definecolor{cb1base}{HTML}{A7ADE6}%
\definecolor{cb2base}{HTML}{97C0E0}
\definecolor{cb3base}{HTML}{8BD0DC}
\definecolor{cb4base}{HTML}{8DDEC6}
\definecolor{cb5base}{HTML}{8FEBB2}
\colorlet{cb0}{cb0base!78!white}
\colorlet{cb1}{cb1base!78!white}
\colorlet{cb2}{cb2base!78!white}
\colorlet{cb3}{cb3base!78!white}
\colorlet{cb4}{cb4base!78!white}
\colorlet{cb5}{cb5base!78!white}

\definecolor{wincol}{HTML}{1B4F72}   %
\definecolor{losscol}{HTML}{7B241C}  %
\colorlet{wincolbg}{cb5base!62!white}
\colorlet{losscolbg}{cb0base!62!white}

\usepackage{amsmath}       %
\usepackage{amssymb}       %
\usepackage{amsfonts}      %
\usepackage{mathtools}     %
\usepackage{bm}            %
\usepackage{bbm}
\usepackage{amsmath}
\usepackage{amssymb}

\usepackage{cleveref}       %

\usepackage{xspace}            %
\usepackage{microtype}         %
\usepackage{afterpage}         %
\usepackage{dblfloatfix}       %
\usepackage{cuted}             %
\usepackage{placeins}          %
\usepackage{enumitem}          %
\usepackage{booktabs}          %
\usepackage{algorithm}         %
\usepackage{algpseudocode}     %
\usepackage{dsfont}

\usepackage{comment}

\usepackage{amsthm}

\newtheorem{definition}{Definition}

\newcommand{\R}{\mathbb{R}}    %

\renewcommand{\v}{\bm{v}}  %
\renewcommand{\a}{\mathbf{a}}
\renewcommand{\b}{\mathbf{b}}  %
\renewcommand{\c}{\mathbf{c}}

\newcommand{\W}{\mathbf{W}}
\newcommand{\e}{\mathbf{e}}
\newcommand{\w}{\mathbf{w}}

\newcommand{\alphaa}{\bm{\alpha}}
\newcommand{\betaa}{\bm{\beta}}

\newcommand{\deltaa}{\bm{\delta}}
\newcommand{\thetaa}{\bm{\theta}}
\newcommand{\x}{\mathbf{x}}

\newcommand{\sigmaa}{\bm{\sigma}}
\newcommand{\muu}{\bm{\mu}}
\newcommand{\s}{\mathbf{s}}

\newcommand{\indicator}{\mathds{1}}

\usepackage{caption}
\DeclareCaptionFont{appendixten}{\fontsize{10pt}{12pt}\selectfont}
\setlist{nosep}

\newcommand{\parhead}[1]{\noindent\textbf{#1}\hspace{0.5em}}

%% file: sections/1-introduction.tex
\section{Introduction}
\label{sec:introduction}

\begin{figure*}[!ht]
  \centering
    \includegraphics[width=\linewidth]{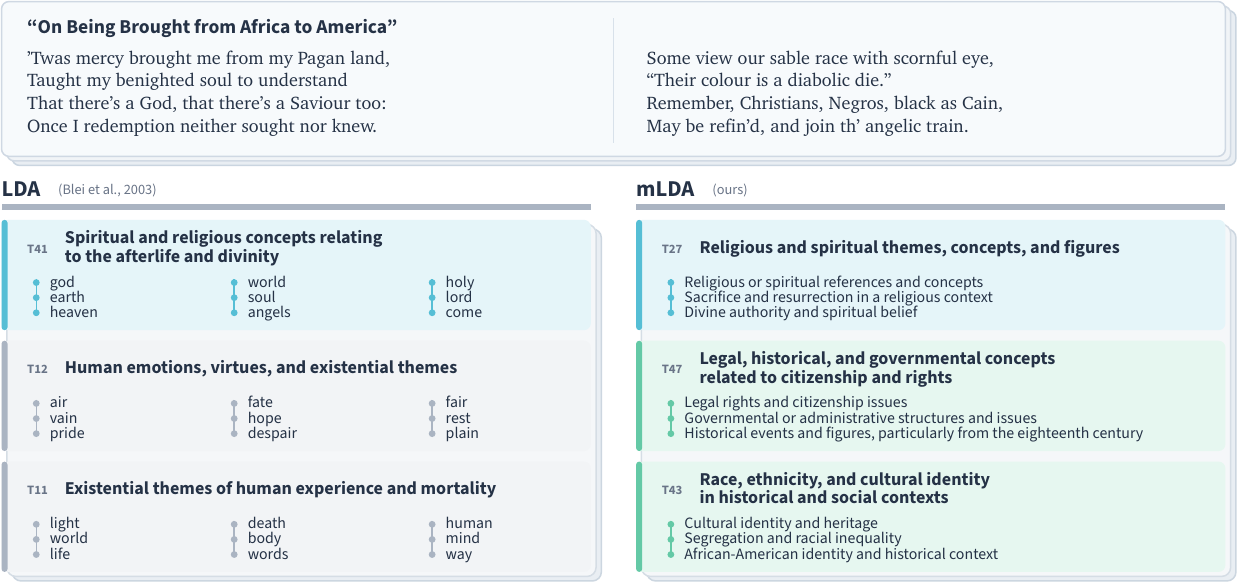}
  \caption{
Comparison of LDA and mLDA topic outputs for Phillis Wheatley's ``On Being Brought from Africa to America'' from PoemSum. The top three topics from each model are shown with LLM-generated summaries. LDA topics are represented by their top words, while mLDA topics are represented by selected top SAE features---natural-language descriptions of learned directions in LLM activation space. Both models recover the poem's religious aspect (blue). LDA's remaining topics are relatively broad poetic themes (gray); mLDA instead identifies the poem's historical and racial context (green).
}
\label{fig:figure1}
\end{figure*}

Topic models are a family of unsupervised algorithms that automatically discover thematic structure in document collections \citep{blei2012probabilistic}. Given a corpus of texts, they produce a predefined number of topics—each represented by a set of words that characterize the theme—along with a per-document breakdown indicating how much each topic contributes to that document's content.

Traditional methods, such as Latent Dirichlet Allocation (LDA) \cite{blei2003latent}, represent documents as bag-of-words counts and discover topics by modeling patterns of word co-occurrence. However, by operating on bag-of-words representations, they miss contextual and semantic nuances.
Neural topic models \citep[e.g.,][]{bianchi-etal-2021-pre, grootendorst2022bertopic, wu2024fastopic} attempt to mitigate this limitation by leveraging pretrained embeddings to capture richer semantics. However, these models interpret topics as lists of words weighted by importance,
which restricts their ability to articulate abstract concepts and nuanced semantic relationships.
Even when probabilistic neural topic models incorporate pretrained embeddings, they still model the generation of word counts, implicitly retaining bag-of-words assumptions.

Independently, recent advances in \textit{mechanistic interpretability} have shown that many high-level semantic concepts in large language models (LLMs) are encoded as linear directions within their internal activations \citep{mikolov2013linguistic, elhage2022toy}.
Sparse autoencoders (SAEs) \citep{tamkin2023codebook, cunningham2023sparse} are neural models that extract these interpretable features from LLM activations, each of which can be subsequently labeled with automatically generated textual descriptions \citep{bills2023language,paulo2024automaticallyinterpreting}.

In this paper, we explore the use of these features for topic modeling.
Unlike traditional bag-of-words
representations, SAE features capture contextual and semantic concepts that extend beyond word co-occurrence patterns. Moreover, since these features can be labeled with textual descriptions, they enable discovering and describing topics at higher semantic abstraction levels.

We introduce Mechanistic Topic Models (MTMs), a family of topic models that adapt existing approaches to operate on SAE features rather than words. This adaptation enables MTMs to: (1) capture context and semantic nuance using pretrained LLM representations;
(2) generate interpretable topic descriptions using SAE features that directly capture abstract concepts like style, tone, and discourse patterns; and (3) enable topic-based controlled generation through learned steering vectors.
\Cref{fig:figure1} compares LDA and mLDA topic representations for a poem by Phillis Wheatley. Both models assign a religious topic; mLDA additionally assigns one topic on historical and civic concepts and another on race and cultural identity, while LDA's other topics are comparatively broad. This example illustrates how MTMs can represent abstract concepts that are difficult to capture with word lists alone.

We instantiate MTMs by adapting three standard topic models to SAEs: mechanistic LDA (mLDA) from \citet{blei2003latent}, mechanistic Embedded Topic Model (mETM) from \citet{dieng2020topic}, and mechanistic BERTopic (mBERTopic) from \citet{grootendorst2022bertopic}.

We make three contributions. First, we introduce Mechanistic Topic Models (MTMs) alongside three variants and demonstrate that these models perform well on challenging corpora including abstract texts and short, context-limited documents. Second, we develop \textit{topic judge}, a new evaluation method that uses LLM-based pairwise comparisons to assess how well topics describe documents, enabling fair cross-vocabulary evaluation while capturing semantic nuance. Third, we demonstrate that MTMs enable controllable text generation through topic-based steering vectors without sacrificing generation quality.
These contributions demonstrate MTMs' utility as a new approach to topic modeling with distinct advantages on abstract and short-text datasets, and provide a case study in using interpretability tools for downstream tasks beyond model analysis.

%% file: sections/2-related_work.tex
\section{Related Work}
\label{sec:related_work}

\parhead{Mechanistic interpretability.}
We build on work establishing that many high-level concepts in large language models (LLMs) are encoded as recoverable linear directions \citep{mikolov2013linguistic, elhage2022toy, park2023linear}, and that dictionary learning methods such as SAEs can extract these directions at scale \citep{yun2021transformer, bricken2023monosemanticity, tamkin2023codebook, templeton2024scaling, cunningham2023sparse, gao2024scaling}.
Prior applications have primarily focused on model interpretability and control—using extracted directions as steering vectors or for ablation \citep{panickssery2023steering, turner2023steering, tan2024analysing, arditi2024refusal}—with applications in refusal mitigation, enhancing truthfulness, reasoning correction, and style transfer, among others \citep{sakarvadia2023memory, hernandez2023inspecting, arditi2024refusal, o2024steering, cao2024personalized, wang2025adaptive}.
We extend this line of research by applying SAE features beyond their original interpretability contexts and demonstrate their usefulness for discovering topics.
Recent work has identified some limitations to SAEs, such as underperformance on downstream tasks \citep{smith2025negativesaes, wu2025axbenchsteeringllmssimple} and challenges to the linear representation of concepts in LLMs \citep{engels2025languagemodelfeaturesonedimensionally}. However, these concerns are less critical for our topic modeling application, which uses SAEs for semantic featurization and requires only that some high-level features are represented linearly.

\parhead{Neural topic models.}
Neural topic models address limitations of purely probabilistic approaches \cite{blei2003latent}. They generally fall into three distinct paradigms. The first paradigm involves probabilistic models aiming to reconstruct a word count matrix, often augmented with pretrained embeddings \citep{burkhardt2019decoupling,dieng2020topic,bianchi-etal-2021-pre,bianchi-etal-2021-cross,wu2024fastopic}. The second paradigm frames topic discovery as a clustering task, leveraging embeddings derived from pretrained neural models \citep{angelov2020top2vecdistributedrepresentationstopics, grootendorst2022bertopic, zhang-etal-2022-neural-clustering}. The third paradigm employs LLMs directly, using prompt-based techniques to aggregate or define topics \citep{pham2024topicgpt}.
Mechanistic Topic Models (MTMs) extend the first two paradigms by using SAEs instead of standard embeddings; concurrent work also connects SAEs to topic modeling by viewing the SAE objective itself as MAP inference in a continuous topic model \citep{girrbach2026sparse}. For MTMs, SAE features enable richer, context-aware topic descriptions and allow for controlled text generation through learned steering vectors. We do not directly compare MTMs against third-paradigm models such as TopicGPT \citep{pham2024topicgpt}; despite strong performance, their training requires many costly LLM API calls.

\parhead{Topic model evaluation.}
Automated and human coherence metrics \citep{chang2009reading, newman2010automatic, lau2014machine} have long been the standard for topic model evaluation, but are known to be imperfect proxies for human preferences \citep{hoyle2021automated, doogan2022topic}. Recent work has further shown that coherence metrics can miss important aspects of topic quality, such as document-space organization \citep{pereira2025topiceval}, hierarchical consistency \citep{viegas2025cluhtm}, and the relationship between documents and the topics they are assigned to \citep{pereira2025topiceval}. Topic judge is motivated by similar concerns: it goes beyond coherence by directly evaluating how well topics describe their assigned documents, providing a closer proxy for human preferences compared to topic quality.
Recently, LLMs have demonstrated effectiveness as scalable evaluators across diverse language tasks \citep{naismith2023automated, chiang2023can, stammbach-etal-2023, li2023exploring}.
Pairwise preference rankings by LLMs have proven particularly useful in contexts where relative comparisons are straightforward but eliciting global rankings or pointwise scores is challenging, such as in chatbot evaluation \citep{zheng2023judging, li2024llms, liu2024aligning, chiang2024chatbot, gao2024re}.
Building on these insights, we introduce a tournament-style evaluation framework that leverages pairwise LLM judgments to systematically compare topic models.

%% file: sections/3-background.tex
\section{Background} \label{sec:background}

The \textit{linear representation hypothesis} \citep{mikolov2013linguistic, elhage2022toy, park2023linear, costa2025flathierarchicalextractingsparse}
suggests that  LLMs encode many high-level features as linear directions in their activation spaces. It can be formalized as follows:\footnote{\hspace{0.12em}Several formalizations exist, but we adopt a definition close to that of \citet{costa2025flathierarchicalextractingsparse}, as we think it adheres closely to its conventional usage.}

\begin{definition} \textbf{Linear Representation Hypothesis (LRH):} Any activation vector $\a \in \mathbb{R}^{H}$, where $H$ is the hidden dimension of the transformer, produced at a given layer can be decomposed as \begin{equation} \label{eq:linear_representation_hypothesis} \a = \sum_{i=1}^W \alpha_i \w_i + \b , \end{equation} where 
\begin{itemize} 
\item $\b$ is an input-independent constant vector, 
\item the set $\{ \w_1, \w_2, \ldots, \w_W \}$ consists of nearly orthogonal unit vectors (i.e., $\|\w_i\|_2 = 1$ and $|\langle \w_i, \w_j \rangle| < \epsilon$ for $i \neq j$, with $\epsilon$ being a small positive constant), \item each vector $\w_i$ corresponds to a human-interpretable feature (e.g., semantic content, syntactic structure, or style), \item each scalar $\alpha_i$ represents the strength of feature $i$ in the activation vector $\a$, with sparse activation (i.e., $|\{i : \alpha_i \neq 0 \} | \ll W$), \item the number of vectors $W$ is typically much larger than their dimension $H$. \end{itemize} \end{definition}

This decomposition implies that (i) high-level semantic features that LLMs extract from text can be recovered from model activations, and (ii) we can construct \textit{steering vectors}
$\s = \sum_{i=1}^W \delta_i \w_i$
that when added to $\a = \sum_{i=1}^W \alpha_i \w_i  + \b$,
are equivalent to modifying its feature strengths
$\a + \s = \sum_{i=1}^W (\alpha_i+\delta_i) \w_i  + \b$.
They can be used in generating text to modulate the expression of particular features,
by setting $\delta_i >0 $ to increase the  expression of feature $i$ and $\delta_i <0$ to decrease it.

To identify the LRH feature directions $\{\mathbf{w}_i\}_{i=1}^W$, we can train \textit{sparse autoencoders} (SAEs)---unsupervised models that learn to reconstruct LLM activations. SAEs are parameterized by a single-layer neural network,
\begin{align}
    \alphaa(\a) &= \sigma(\W_{\text{in}} \a + \b_{\text{in}}), \\
    \hat{\a}(\a) &= \W_{\text{out}} \alphaa(\a) + \b_{\text{out}},
\end{align}
where $\sigma$ is an activation function (e.g., ReLU, JumpReLU), and the learnable parameters are $\W_{\text{in}} \in \mathbb{R}^{W \times H}$,  $\W_{\text{out}} \in \mathbb{R}^{H \times W}$, \(\b_{\mathrm{in}}\in\R^W\), and 	\(\b_{\mathrm{out}}\in\R^H\). The network is trained to minimize a regularized reconstruction error,
\begin{align}
    \mathcal{L}(\a) &= \frac{1}{2} \|\hat{\a}(\a) - \a\|_2^2 + \lambda \|\alphaa(\a)\|_0.
\end{align}

Once trained, the feature directions $\{\w_i\}_{i=1}^W$ are identified with the columns of $\mathbf{W}_{\text{out}}$. The encoder network $\alphaa(\a)$ can also be used to decompose any activation onto the feature directions and obtain its feature activation strengths $\alpha_i$.

After training, the learned feature directions can be automatically interpreted into textual descriptions $\{\texttt{d}_i\}_{i=1}^W$ through computational methods \citep{bills2023language,paulo2024automaticallyinterpreting,templeton2024scaling}. A common approach involves computing feature activation strengths over a large corpus, selecting tokens with the highest activation strengths along with their surrounding context, and prompting an LLM to produce a short description of the underlying feature \citep{bills2023language}.

%% file: sections/4-method.tex
\section{Mechanistic Topic Models}
\label{sec:methods}

We introduce Mechanistic Topic Models (MTMs), which extend topic modeling by using SAE features.
This shift provides three advantages: (1) semantic richness, as SAE features capture context-aware and semantically abstract concepts; (2)
topic descriptions that can articulate complex themes that are hard to convey through word lists alone; and (3) topic steering vectors that can be used for topic-based controlled generation.

All MTMs share the same workflow. Given a corpus $\mathcal{D}$ of $D$ documents and a desired number of topics $K$, we first transform documents into SAE feature counts (\Cref{sec:methods:featurization}), then learn topic-feature weights $\boldsymbol{\beta}_k \in \mathbb{R}_+^W$ and document-topic distributions $\boldsymbol{\theta}_d \in \Delta^{K-1}$ (\Cref{sec:methods-topic-model instantiations}), generate interpretable topic descriptions $\texttt{t}_k$ from learned features (\Cref{sec:methods:interpreting_topics}), and construct steering vectors $\mathbf{s}_k$ for controllable generation (\Cref{sec:method:steering}). We first describe the featurization process and then detail three specific MTM variants.

\subsection{From Documents to SAE Features}
\label{sec:methods:featurization}

MTMs represent documents as SAE feature counts rather than word counts or raw embeddings.
This presents two challenges.

First, unlike words that either appear or not, SAE features have continuous activations at each token position. For a document $d$ containing $N_\text{tok}$ tokens, let $\mathbf{a}_{d,j} \in \mathbb{R}^H$ denote the LLM hidden-state activation at the chosen layer for token $j$. We use a thresholding approach to count how often each feature $i$ activates strongly across $(\mathbf{a}_{d,1}, \dots, \mathbf{a}_{d,N_\text{tok}})$,
\begin{equation}
\label{eq:count-direction-vectors}
\tilde{c}_{d,i} = \sum_{j=1}^{N_{\text{tok}}} \indicator \{ \alpha_{i}(\mathbf{a}_{d,j}) > q_i\},
\end{equation}
where $\alpha_i(\mathbf{a}_{d,j})$ is feature $i$'s activation on token $j$, and $q_i$ is the 80th percentile of feature $i$'s nonzero activation distribution on the original SAE training data, although any dataset of interest
can be used. This approach produces interpretable counts, adapts to each feature's typical activation range, and prevents activation false positives.

Second, SAEs can learn spurious features with unclear meanings or mislabeled descriptions. We address this through an LLM-based preprocessing step that filters out likely spurious or topic-irrelevant features (e.g., low-level grammatical features), and an optional post-training refinement of topic descriptions. These quality control measures are detailed in Appendix~\ref{app:mtm_implementation_details}.

The feature vectors $\{\tilde{\mathbf{c}}_d\}_{d=1}^D$ serve as input to all MTMs described below. For the rest of the paper, we use $W$ to denote the number of features after this filtering.

\subsection{MTM Variants} \label{sec:methods-topic-model instantiations}
Having transformed documents into SAE feature vectors $\{\tilde{\c}_d\}_{d=1}^D$, we now apply topic modeling algorithms to these representations. %

We propose three variants with different adaptation strategies: mechanistic LDA (mLDA) provides a straightforward extension of LDA \citep{blei2003latent} by treating features as words; mechanistic ETM (mETM) adapts ETM by representing topics as LLM activation vectors \citep{dieng2020topic}; and mechanistic BERTopic (mBERTopic) takes a clustering approach using SAE feature directions to construct document embeddings \citep{grootendorst2022bertopic}.

\subsubsection{Mechanistic LDA (mLDA)}
\label{sec:methods:mlda}

Latent Dirichlet Allocation (LDA) \citep{blei2003latent} is a foundational probabilistic topic model that represents documents as mixtures of topics and topics as distributions over words. Mechanistic LDA (mLDA) adapts this model by replacing the topic-word distributions with distributions over SAE features. Following LDA's generative process, we assume
\begin{align}
    \boldsymbol{\betaa}_k &\sim \text{Dirichlet}_W(\eta),  \\
    \boldsymbol{\thetaa}_d &\sim \text{Dirichlet}_K(\alpha),
\end{align}
where $\betaa_k \in \Delta^{W-1}$ is a distribution over the $W$ feature directions learned by the SAE and $\thetaa_d \in \Delta^{K-1}$ is the document distribution over $K$ topics.

The SAE feature counts are generated by
\begin{equation}
    \tilde{\c}_d \sim \text{Multinomial}(\boldsymbol{\thetaa}_d \boldsymbol{\betaa}, N_{\text{sae}}),
\end{equation}
where $N_{\text{sae}} = \sum_i \tilde{c}_{d,i}$ is the total SAE feature count in the document. While the multinomial assumption has limitations (\Cref{sec:methods:metm}), here we retain it to leverage existing LDA inference algorithms. Depending on the dataset size, we use standard collapsed Gibbs sampling \citep{griffiths2004finding} or variational inference \citep{blei2003latent} to approximate posterior distributions over $\{\betaa_k\}$ and $\{\thetaa_d\}$ (see \Cref{app:baseline_implementations}).

\subsubsection{Mechanistic ETM (mETM)}
\label{sec:methods:metm}

The Embedded Topic Model (ETM) \citep{dieng2020topic} represents topics as vectors in word embedding space, leveraging these word embeddings to capture semantic relationships. Mechanistic ETM (mETM) extends this idea to the space of LLM activations, representing each topic $k$ as a learned LLM activation $\v_k \in \mathbb{R}^H$.

The generative process first samples document-topic proportions from a logistic-normal, %
\begin{align}
    \boldsymbol{\deltaa}_d &\sim \mathcal{N}(\mathbf{0}, \mathbf{I}), \\
    \boldsymbol{\thetaa}_d &= \text{softmax}(\boldsymbol{\deltaa}_d).
\end{align}

Each topic-feature distribution $\betaa_k \in [0, 1]^W$ is obtained by transforming the learned activation $\v_k$ as in the SAE encoder,
\begin{equation}
\label{eq:metm-sae}
\betaa_k = \sigma\left( \W_{\text{in}} \v_k + \b\right),
\end{equation}
where $\W_{\text{in}}$ is initialized with the SAE encoder matrix and fixed during training, $\b$ is a learned bias vector, and $\sigma$ is the sigmoid function. Here, $\beta_{k,i}$ represents the probability that feature $i$ is active in a document token from topic $k$.

Unlike in mLDA, feature counts are drawn from a binomial distribution
\begin{equation}
\label{eq:c-sampling-metm}
\tilde{c}_{d,i} \sim \text{Binomial}([\thetaa_d \betaa]_i, N_{\text{tok}}).
\end{equation}
This distribution is chosen to respect the constraint that each feature activates at most once per token.
Following \citet{dieng2020topic}, we use amortized variational inference with neural networks $\muu_\phi$ and $\sigmaa_\phi$ to parameterize the variational distribution $q_\phi(\deltaa_d) = \mathcal{N}(\muu_\phi(\tilde{\c}_d), \sigmaa_\phi(\tilde{\c}_d))$. The parameters $\phi$, $\{\v_k\}$, and $\b$ are jointly optimized by maximizing the ELBO,
\begin{align}
\sum_{d=1}^D
\mathbb{E}_{q_\phi}
\left[
\log
\frac{
p(\tilde{\c}_d \mid \deltaa_d, \{\v_k\}, \b)\,
p(\deltaa_d)
}{
q_\phi(\deltaa_d \mid \tilde{\c}_d)
}
\right].
\end{align}

\subsubsection{Mechanistic BERTopic (mBERTopic)}
\label{sec:methods:mbertopic}

BERTopic \citep{grootendorst2022bertopic} frames topic modeling as a clustering problem in document embedding space. Mechanistic BERTopic (mBERTopic) takes a similar approach but forms document embeddings $\tilde{\e}_d$ from the SAE feature representation,
\begin{equation}
\label{eq:mbertopic}
\tilde{\e}_d = \frac{1}{N_{\text{tok}}} \sum_{i=1}^W \tilde{c}_{d,i} \w_i,
\end{equation}
where $\w_i$ is the $i^\text{th}$ feature direction from the SAE decoder matrix.

Following \citet{grootendorst2022bertopic}, we apply UMAP \citep{mcinnes2018umap} for dimensionality reduction followed by HDBSCAN clustering \citep{mcinnes2017hdbscan}  to the document embeddings $\{\tilde{\e}_d\}$. We then extract topic-feature distributions using class-based TF-IDF, which treats each cluster as a meta-document,
\begin{equation}
\beta_{k,i} \propto \text{tf}_{k,i} \cdot \log\left(1 + \frac{A}{\text{tf}_i}\right),
\end{equation}
where $\text{tf}_{k,i} = \sum_{d \in \mathcal{D}_k} \tilde{c}_{d,i}$ is the count of feature $i$ across all documents in cluster $k$, $A$ is the average count of all features per cluster, and $\text{tf}_i$ is the total count of feature $i$ across all clusters.

We experimented with alternative embeddings in \Cref{eq:mbertopic}, including using the SAE activations as opposed to counts, and skipping the pre-filtering step detailed in \Cref{sec:methods:featurization}.
However, the formulation in \Cref{eq:mbertopic} was consistently chosen by our hyperparameter optimization procedures, which we believe affirms the usefulness and importance of the steps detailed in \Cref{sec:methods:featurization}.

\subsection{Using MTMs}
\label{sec:methods:using-mtms}
Once trained, all the MTMs above produce document-topic proportions $\thetaa_d$ and topic-feature distributions $\betaa_k$ that can be used in downstream applications. Of particular significance to MTMs are topic interpretations and steering vectors.

\paragraph{Interpreting Topics.}\label{sec:methods:interpreting_topics}
Topic interpretation in MTMs follows an approach similar to conventional topic models. Given a learned topic $k$, we identify the top $n$ SAE features by selecting those with the highest weights in the topic-feature vector $\betaa_k$. We then construct the textual topic representation $\texttt{t}_k$ from the automatically generated descriptions $\texttt{d}_i$ corresponding to these $n$ features.

The last step can be done in two ways. The TopFeatures (TF) approach directly uses the feature descriptions $\texttt{d}_i$
and concatenates them. The Summarization approach (Sum.) further processes the concatenated text by passing it through an LLM to convert it into a one-sentence summary. Summarization is beneficial, as it can also be applied to word-based models for standardization in evaluations. \Cref{fig:figure1} shows both approaches in use. \Cref{app:all_prompts:topic_summarization} provides the summarization prompt
used in this paper.

\paragraph{Steering.}\label{sec:method:steering}
One advantage of MTMs is their ability to steer text generation toward discovered topics. We achieve this by constructing a topic-specific steering vector $\s_k$ that we add to the LLM's activations to bias generation toward topic $k$.

Each steering vector $\s_k$ is constructed by weighting SAE feature directions $\{\w_i\}$ according to their importance in topic $k$ as captured by the topic-feature weights $\betaa_k \in \mathbb{R}_+^W$,
\begin{equation}
\s_k = \frac{\sum_{i \in W} \betaa_{k,i}\w_i}{\left\| \sum_{i \in W} \betaa_{k,i}\w_i\right\|_2}.
\label{eq:steering_vector}
\end{equation}
\Cref{eq:steering_vector} is a unit vector that points in the direction most characteristic of topic $k$ in the LLM's activation space.

 To control the intensity of topic steering, we use an intervention that first removes any existing topic signal before adding the desired amount. Consider centered activations $\bar{\a} = \a - \b$, where $\b$ is the bias from \Cref{eq:linear_representation_hypothesis}. We decompose the activation into components parallel and perpendicular to the steering direction,
\begin{align}
    \bar{\a}_{\parallel} &= (\bar{\a} \cdot \s_k) \s_k, \\
    \bar{\a}_{\perp} &= \bar{\a} - \bar{\a}_\parallel.
\end{align}
The steered activation replaces the parallel component with a scaled steering vector,
\begin{equation}
    \label{eq:steered:interpretable}
    \a_{\text{steered}} = \bar{\a}_{\perp} + \lambda \s_k + \b,
\end{equation}
where $\lambda$ controls the steering strength, allowing for modulation of topic expression in generated text.

For example, one specific topic learned by mLDA on the Bills dataset (\Cref{sec:empirical_studies}) places high weights on SAE features with descriptions: ``specific legal terms and conditions related to immigration status'', ``references to government policies and legal regulations'', and ``references to labor conditions and economic structures''. We then form a steering vector $\s_k$ from this topic using \Cref{eq:steering_vector}. Applying this steering vector to the prompt ``A text about'' results in the continuation `` a person who is not of the United States, but has been granted permission to enter the country. The term `temporary resident' (TR) refers to people who have entered the U.S. and are allowed to stay in the US\dots'', an example of successful topic-guided text generation.

\begin{table*}[!ht]
  \footnotesize
  \centering
  \begin{minipage}[t]{0.723\textwidth}
    \vspace{0pt}
    \centering
    \fontsize{9.3pt}{9.8pt}\selectfont
    \setlength{\tabcolsep}{2.75pt}
    \renewcommand{\arraystretch}{1.10}
    \setlength{\aboverulesep}{0.50ex}
    \setlength{\belowrulesep}{0.50ex}
    \newcommand{\datasetsep}{\arrayrulecolor{gray!45}\specialrule{\lightrulewidth}{0.16ex}{0.16ex}\arrayrulecolor{black}}
      \begin{tabular*}{\linewidth}{@{\extracolsep{\fill}}ll rr rr rr rr@{}}
        \toprule
        & & & & \multicolumn{2}{c}{LDA} & \multicolumn{2}{c}{ETM} & \multicolumn{2}{c}{BERTopic} \\
        \cmidrule(lr){5-6} \cmidrule(lr){7-8} \cmidrule(lr){9-10}
        Data & & D-VAE & FAST
        & Base & MTM & Base & MTM & Base & MTM \\
        \midrule
        \multirow{2}{*}{20NG}
          & TF   & 1094 & 1419
            & \cellcolor{losscolbg}\underline{1580}          & \cellcolor{wincolbg}\bfseries1613
            & \cellcolor{wincolbg}\underline{1602}     & \cellcolor{losscolbg}\underline{1601}
            & \cellcolor{losscolbg}1544                      & \cellcolor{wincolbg}1548 \\
          & Sum. & 1322 & 1470
            & \cellcolor{losscolbg}1514                      & \cellcolor{wincolbg}\bfseries1614
            & \cellcolor{losscolbg}1510                      & \cellcolor{wincolbg}1573
            & \cellcolor{losscolbg}1484                      & \cellcolor{wincolbg}1514 \\
        \datasetsep
        \multirow{2}{*}{Bills}
          & TF   & 1151 & 1316
            & \cellcolor{wincolbg}\bfseries1763        & \cellcolor{losscolbg}1571
            & \cellcolor{wincolbg}1670                 & \cellcolor{losscolbg}1584
            & \cellcolor{wincolbg}1509                       & \cellcolor{losscolbg}1436 \\
          & Sum. & 1395 & 1447
            & \cellcolor{losscolbg}1578                      & \cellcolor{wincolbg}\bfseries1638
            & \cellcolor{losscolbg}1536                      & \cellcolor{wincolbg}\underline{1630}
            & \cellcolor{wincolbg}1410                       & \cellcolor{losscolbg}1366 \\
        \datasetsep
        \multirow{2}{*}{Wiki}
          & TF   & 1304 & 1288
            & \cellcolor{losscolbg}1575                      & \cellcolor{wincolbg}\underline{1641}
            & \cellcolor{losscolbg}1559                      & \cellcolor{wincolbg}\bfseries1641
            & \cellcolor{losscolbg}1443                      & \cellcolor{wincolbg}1549 \\
          & Sum. & 1425 & 1424
            & \cellcolor{losscolbg}1551                      & \cellcolor{wincolbg}\bfseries1628
            & \cellcolor{losscolbg}1494                      & \cellcolor{wincolbg}\underline{1620}
            & \cellcolor{losscolbg}1387                      & \cellcolor{wincolbg}1471 \\
        \datasetsep
        \multirow{2}{*}{Yelp}
          & TF   & 1095 & 1271
            & \cellcolor{losscolbg}1644                      & \cellcolor{wincolbg}\bfseries1715
            & \cellcolor{losscolbg}1558                      & \cellcolor{wincolbg}\underline{1675}
            & \cellcolor{losscolbg}1430                      & \cellcolor{wincolbg}1611 \\
          & Sum. & 1342 & 1383
            & \cellcolor{wincolbg}\bfseries1627        & \cellcolor{losscolbg}\underline{1621}
            & \cellcolor{losscolbg}1488                      & \cellcolor{wincolbg}\underline{1609}
            & \cellcolor{losscolbg}1347                      & \cellcolor{wincolbg}1585 \\
        \datasetsep
        \multirow{2}{*}{AGN.}
          & TF   & 1402 & 1462
            & \cellcolor{losscolbg}\underline{1552}    & \cellcolor{wincolbg}\bfseries1576
            & \cellcolor{losscolbg}1511                      & \cellcolor{wincolbg}1537
            & \cellcolor{losscolbg}1476                      & \cellcolor{wincolbg}1484 \\
          & Sum. & 1480 & 1491
            & \cellcolor{losscolbg}1489                      & \cellcolor{wincolbg}\underline{1590}
            & \cellcolor{losscolbg}1431                      & \cellcolor{wincolbg}\bfseries1607
            & \cellcolor{losscolbg}1442                      & \cellcolor{wincolbg}1469 \\
        \midrule
        \multirow{2}{*}{GoEmo.}
          & TF   & 1185 & 1431
            & \cellcolor{losscolbg}1442                      & \cellcolor{wincolbg}\bfseries1728
            & \cellcolor{losscolbg}1444                      & \cellcolor{wincolbg}1664
            & \cellcolor{losscolbg}1428                      & \cellcolor{wincolbg}1678 \\
          & Sum. & 1346 & 1419
            & \cellcolor{losscolbg}1452                      & \cellcolor{wincolbg}\bfseries1630
            & \cellcolor{losscolbg}1449                      & \cellcolor{wincolbg}\underline{1630}
            & \cellcolor{losscolbg}1463                      & \cellcolor{wincolbg}\underline{1611} \\
        \datasetsep
        \multirow{2}{*}{Poem.}
          & TF   & 1185 & 1306
            & \cellcolor{losscolbg}1573                      & \cellcolor{wincolbg}\underline{1662}
            & \cellcolor{losscolbg}1524                      & \cellcolor{wincolbg}\bfseries1678
            & \cellcolor{losscolbg}1453                      & \cellcolor{wincolbg}1619 \\
          & Sum. & 1360 & 1412
            & \cellcolor{losscolbg}1522                      & \cellcolor{wincolbg}1536
            & \cellcolor{losscolbg}1446                      & \cellcolor{wincolbg}\underline{1597}
            & \cellcolor{losscolbg}1530                      & \cellcolor{wincolbg}\bfseries1598 \\
        \datasetsep
        \multirow{2}{*}{WriPro.}
          & TF   & 918 & 1263
            & \cellcolor{losscolbg}1545                      & \cellcolor{wincolbg}\bfseries1813
            & \cellcolor{losscolbg}1538                      & \cellcolor{wincolbg}\underline{1808}
            & \cellcolor{losscolbg}1385                      & \cellcolor{wincolbg}1731 \\
          & Sum. & 1373 & 1375
            & \cellcolor{losscolbg}1439                      & \cellcolor{wincolbg}\underline{1674}
            & \cellcolor{losscolbg}1430                      & \cellcolor{wincolbg}\bfseries1678
            & \cellcolor{losscolbg}1394                      & \cellcolor{wincolbg}1637 \\
        \bottomrule
      \end{tabular*}%
  \end{minipage}\hfill
  \begin{minipage}[t]{0.264\textwidth}
    \vspace{0pt}
    \centering
    \includegraphics[width=\linewidth]{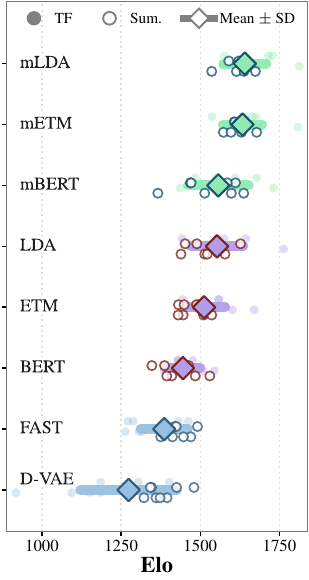}
  \end{minipage}
  \caption{Topic judge Elo scores across datasets, with a model-level summary at right. The first five datasets are standard benchmark datasets; the last three are hard (short and/or more abstract). \textbf{Bold} indicates the row best (bootstrap confidence interval, $p < 0.05$); underlining indicates $p > 0.05$ against bold. Green \textcolor{cb5}{\rule{0.6em}{0.6em}} indicates an equal or higher score than the corresponding word-based model, and purple \textcolor{cb0}{\rule{0.6em}{0.6em}} a lower score.}
  \label{tab:topic_judge_standard}
  \label{tab:topic_judge_hard}
\end{table*}

%% file: sections/5-empirical_studies.tex
\begin{figure*}[t]
  \centering
  \includegraphics[width=\textwidth]{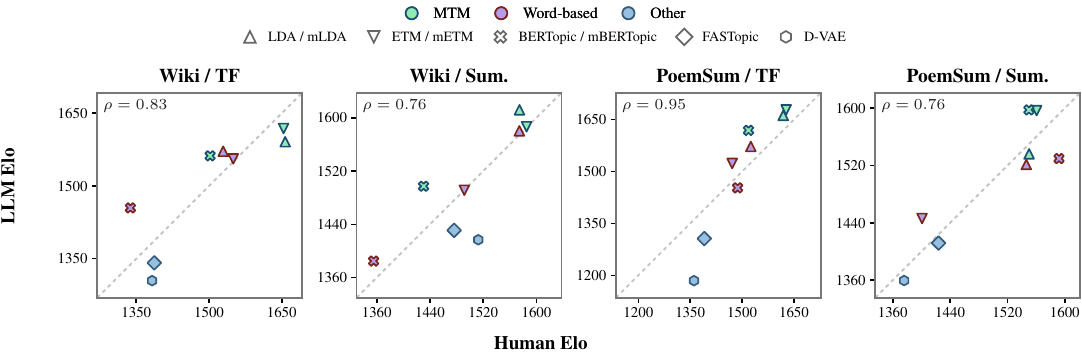}
  \caption{Human vs.\ LLM topic-judge Elo scores on Wiki and PoemSum. Each panel compares the Elo scores assigned by the human study and the GPT-4.1 judge for one dataset/representation condition. The dashed diagonal indicates perfect agreement, and each panel reports Spearman's $\rho$. The panels and corresponding $\rho$ values show strong agreement between human and LLM Elo scores.}
  \label{fig:human_llm_elo}
\end{figure*}

\section{Empirical Studies}
\label{sec:empirical_studies}

We evaluate MTMs and baselines using standard topic modeling metrics (coherence, topic diversity, and alignment) 
and a novel metric, topic judge, which performs a series of comparisons between pairs of models, asking an LLM judge which set of topics they prefer with respect to a reference document from the corpus. We find that an MTM is preferred by topic judge in most datasets, while MTMs are largely comparable to baselines on standard metrics, with particular strengths on abstract and short-text datasets. Finally, we show that MTMs capture novel topics, and that their topics can be used to steer LLM generations.

\paragraph{Datasets.} We study eight datasets spanning a range of domains, document lengths, and thematic characteristics: online newsgroup posts (20NG; \citealt{lang1995newsweeder}), bill summaries from the 110--114$^\text{th}$ U.S. congresses (Bills; \citealt{adler2011congressional, hoyle-etal-2022-neural}), Wikipedia articles labeled as ``good'' by editors (Wiki; \citealt{merity-etal-2017-pointer}), short Reddit comments expressing emotion (GoEmotions; \citealt{demszky-etal-2020-goemotions}), a collection of poems (PoemSum; \citealt{mahbub-etal-2023-poemsum}), news articles across four categories (AGNews; \citealt{zhang2015character}), business reviews (Yelp Polarity; \citealt{zhang2015character}), and creative writing stories (WritingPrompts; \citealt{fan-etal-2018-hierarchical}).

We expected GoEmotions to be more challenging due to its short documents, and
PoemSum and WritingPrompts to be more challenging due to their abstract themes.

\Cref{app:datasets} contains dataset statistics, preprocessing steps for word-based models, and information on labels used for the topical alignment metric.

\paragraph{Models.} We compare to the word-based counterparts of MTMs: LDA \citep{blei2003latent}, ETM \citep{dieng2020topic}, and BERTopic \citep{grootendorst2022bertopic}. We also compare to two other neural models: Dirichlet VAE (D-VAE) \citep{burkhardt2019decoupling}, which is a VAE product-of-experts model; and FASTopic \citep{wu2024fastopic}, which is a model using optimal transport alongside pretrained embeddings.

\paragraph{Setup.} For MTMs, we use the GemmaScope family of SAEs trained on Gemma 2-9B LLM activations \citep{team2024gemma, lieberum2024gemmascopeopensparse}. We use the 16k SAE from layer 10 throughout, as we found empirically that it provides the most useful features for topic modeling, though practitioners may prefer other layers depending on the characteristics of the dataset. The SAE feature metadata, including descriptions, are downloaded from Neuronpedia \citep{neuronpedia}. The implementation details for the baseline models are in \Cref{app:baseline_implementations}.

For our experiments, we choose the number of topics to be $K=50$ for AGNews, GoEmotions, and PoemSum, as these are the smallest datasets, and $K=100$ for the remaining datasets. To select model hyperparameters, we use Bayesian optimization and optimize the topic quality metric proposed in \citet{dieng2020topic} for each model-dataset pair. \Cref{app:bayesian_optimization} contains details on this procedure.

\afterpage{%
\afterpage{%
\begin{table*}[t]
  \centering
  \resizebox{\textwidth}{!}{%
  \small
  \renewcommand{\arraystretch}{1.1}
      \begin{tabular}{@{}l cc cc cc cc cc cc cc cc@{}}
        \toprule
        & \multicolumn{2}{c}{20NG} & \multicolumn{2}{c}{Bills} & \multicolumn{2}{c}{Wiki} & \multicolumn{2}{c}{Yelp} & \multicolumn{2}{c}{AGNews} & \multicolumn{2}{c}{GoEmo.} & \multicolumn{2}{c}{PoemSum} & \multicolumn{2}{c}{WriPro.} \\
        \cmidrule(lr){2-3} \cmidrule(lr){4-5} \cmidrule(lr){6-7} \cmidrule(lr){8-9} \cmidrule(lr){10-11} \cmidrule(lr){12-13} \cmidrule(lr){14-15} \cmidrule(lr){16-17}
        Model
          & R & I
          & R & I
          & R & I
          & R & I
          & R & I
          & R & I
          & R & I
          & R & I \\
        \midrule
        D-VAE
          & \cellcolor{cb1} 2.09 & \cellcolor{cb1} 0.55
          & \cellcolor{cb3} 2.58 & \cellcolor{cb2} 0.66
          & \cellcolor{cb5} 2.93 & \cellcolor{cb5} \bfseries 0.91
          & \cellcolor{cb0} 1.75 & \cellcolor{cb0} 0.46
          & \cellcolor{cb4} 2.79 & \cellcolor{cb4} 0.81
          & \cellcolor{cb0} 1.77 & \cellcolor{cb0} 0.38
          & \cellcolor{cb0} 1.76 & \cellcolor{cb0} 0.37
          & \cellcolor{cb0} 1.77 & \cellcolor{cb0} 0.44 \\
        FASTopic
          & \cellcolor{cb3} 2.43 & \cellcolor{cb3} 0.71
          & \cellcolor{cb4} 2.70 & \cellcolor{cb3} 0.74
          & \cellcolor{cb5} 2.89 & \cellcolor{cb4} 0.87
          & \cellcolor{cb2} 2.22 & \cellcolor{cb1} 0.59
          & \cellcolor{cb5} 2.82 & \cellcolor{cb4} \underline{0.85}
          & \cellcolor{cb4} 2.70 & \cellcolor{cb4} 0.81
          & \cellcolor{cb0} 1.97 & \cellcolor{cb0} 0.43
          & \cellcolor{cb2} 2.33 & \cellcolor{cb2} 0.66 \\
        \midrule
        LDA
          & \cellcolor{cb3} 2.65 & \cellcolor{cb3} 0.75
          & \cellcolor{cb5} \bfseries 2.96 & \cellcolor{cb3} 0.71
          & \cellcolor{cb5} 2.83 & \cellcolor{cb3} 0.77
          & \cellcolor{cb2} 2.35 & \cellcolor{cb2} 0.61
          & \cellcolor{cb5} 2.83 & \cellcolor{cb3} 0.75
          & \cellcolor{cb2} 2.20 & \cellcolor{cb1} 0.56
          & \cellcolor{cb2} 2.30 & \cellcolor{cb1} 0.55
          & \cellcolor{cb4} 2.63 & \cellcolor{cb3} \underline{0.79} \\
        ETM
          & \cellcolor{cb2} 2.40 & \cellcolor{cb3} 0.74
          & \cellcolor{cb5} 2.81 & \cellcolor{cb3} 0.72
          & \cellcolor{cb4} 2.72 & \cellcolor{cb4} 0.84
          & \cellcolor{cb2} 2.32 & \cellcolor{cb2} 0.65
          & \cellcolor{cb3} 2.43 & \cellcolor{cb2} 0.60
          & \cellcolor{cb0} 1.90 & \cellcolor{cb0} 0.37
          & \cellcolor{cb1} 2.17 & \cellcolor{cb1} 0.51
          & \cellcolor{cb3} 2.55 & \cellcolor{cb4} \underline{0.81} \\
        BERTopic
          & \cellcolor{cb5} \bfseries 2.87 & \cellcolor{cb4} 0.80
          & \cellcolor{cb5} 2.93 & \cellcolor{cb3} 0.78
          & \cellcolor{cb5} \bfseries 2.97 & \cellcolor{cb4} 0.86
          & \cellcolor{cb5} \bfseries 2.84 & \cellcolor{cb3} 0.74
          & \cellcolor{cb5} \bfseries 2.92 & \cellcolor{cb4} 0.81
          & \cellcolor{cb4} \underline{2.76} & \cellcolor{cb3} 0.80
          & \cellcolor{cb3} 2.36 & \cellcolor{cb1} 0.51
          & \cellcolor{cb3} 2.53 & \cellcolor{cb2} 0.65 \\
        \midrule
        MTM (w/ LDA)
          & \cellcolor{cb4} 2.63 & \cellcolor{cb4} \underline{0.83}
          & \cellcolor{cb5} 2.88 & \cellcolor{cb4} \bfseries 0.81
          & \cellcolor{cb4} 2.77 & \cellcolor{cb4} 0.85
          & \cellcolor{cb4} 2.77 & \cellcolor{cb3} \underline{0.78}
          & \cellcolor{cb4} 2.71 & \cellcolor{cb4} 0.80
          & \cellcolor{cb4} \underline{2.77} & \cellcolor{cb3} 0.80
          & \cellcolor{cb4} \bfseries 2.61 & \cellcolor{cb3} \underline{0.73}
          & \cellcolor{cb4} \bfseries 2.78 & \cellcolor{cb4} \underline{0.80} \\
        MTM (w/ ETM)
          & \cellcolor{cb3} 2.57 & \cellcolor{cb4} 0.80
          & \cellcolor{cb5} 2.87 & \cellcolor{cb3} 0.79
          & \cellcolor{cb4} 2.79 & \cellcolor{cb4} 0.85
          & \cellcolor{cb4} 2.60 & \cellcolor{cb3} 0.73
          & \cellcolor{cb3} 2.46 & \cellcolor{cb3} 0.73
          & \cellcolor{cb4} 2.70 & \cellcolor{cb3} 0.76
          & \cellcolor{cb3} 2.54 & \cellcolor{cb3} 0.70
          & \cellcolor{cb4} 2.65 & \cellcolor{cb3} 0.76 \\
        MTM (w/ BERTopic)
          & \cellcolor{cb3} 2.58 & \cellcolor{cb4} \bfseries 0.83
          & \cellcolor{cb4} 2.75 & \cellcolor{cb4} \underline{0.81}
          & \cellcolor{cb4} 2.76 & \cellcolor{cb4} 0.85
          & \cellcolor{cb4} 2.73 & \cellcolor{cb3} \underline{0.77}
          & \cellcolor{cb4} 2.74 & \cellcolor{cb4} \underline{0.83}
          & \cellcolor{cb5} \bfseries 2.80 & \cellcolor{cb4} \bfseries 0.88
          & \cellcolor{cb2} 2.28 & \cellcolor{cb1} 0.53
          & \cellcolor{cb4} 2.61 & \cellcolor{cb3} 0.71 \\
        \bottomrule
      \end{tabular}%
  }%
\caption{Ratings (R) and intrusion (I) across all datasets, averaged over five runs. \textbf{Bold} indicates the column best (two-sided t-test, $p < 0.05$); underlining indicates $p > 0.05$ against bold. Color coding: for ratings, purple \textcolor{cb0}{\rule{0.6em}{0.6em}} ($<2.0$) to green \textcolor{cb5}{\rule{0.6em}{0.6em}} ($>2.8$); for intrusion, purple \textcolor{cb0}{\rule{0.6em}{0.6em}} ($<0.5$) to green \textcolor{cb5}{\rule{0.6em}{0.6em}} in 0.1 increments.}
\label{tab:coherence_td_standard}
\end{table*}
}%
}%

\subsection{Topic Judge}
\label{sec:empirical_studies:topic_judge}

Topic model evaluation is challenging. Existing metrics have limitations: topical alignment \citep{hoyle-etal-2022-neural} requires labeled data for attributes of interest and does not assess topic description quality; coherence metrics like ratings or intrusion \citep{chang2009reading, newman2010automatic} measure feature relatedness in topics but ignore their relationship to documents; and automated metrics like $C_{\text{NPMI}}$ may correlate poorly with human judgments \citep{hoyle2021automated} and cannot compare models with different vocabularies. These limitations motivated us to develop \textit{topic judge}, a new evaluation framework for topic models.

\paragraph{Methodology.}
Topic judge is inspired by Chatbot Arena-style rankings \citep{chiang2024chatbot}. It evaluates topic models using pairwise comparisons, where an LLM judge determines which model's topic assignments better capture a given document's content. The comparison results are then aggregated via a Bradley-Terry model \citep{bradleyterry1952}.
The assumption is that superior topic models should produce topics that are more descriptive of the documents they are assigned to.

In detail, the method works by performing pairwise comparisons between all model pairs $(m, m')$ to score $M$ models. For each of $T$ comparisons per pair, topic judge: (1) randomly samples a document $d$; (2) identifies the set of top topics for each model according to $\thetaa_d$—either the top $q$ topics or all topics with cumulative mass below a threshold $p$, whichever yields fewer topics; (3) creates text representations for each topic set $\texttt{t}_k$ using either approach (TopFeatures or Summarization); and (4) prompts an LLM judge to select which representation better captures the document.

After collecting all pairwise comparisons, topic judge aggregates the results using a Bradley-Terry model to compute final scores. This model assumes that model $m$ defeats model $m'$ with probability $\sigma(\pi_m - \pi_{m'})$, where $\pi_m$ represents model $m$'s strength and $\sigma$ is the sigmoid function. We infer these parameters via maximum likelihood and convert them to Elo scores (normalized to average 1500) for interpretability. Elo scores are a rescaling of the Bradley-Terry strength parameters such that $P(m \succ m') = 1/(1 + 10^{(E_{m'} - E_m)/400})$, where $E_m$ and $E_{m'}$ are the Elo scores of models $m$ and $m'$. 

Topic judge addresses limitations of existing metrics: it evaluates topic-document relationships directly without requiring labeled data, uses pairwise comparisons that capture relative quality differences across different vocabularies (words vs. SAE features), and leverages an LLM judge to assess semantic content directly rather than through proxies like $C_{\text{NPMI}}$. For the specific prompt and examples with judge responses, see \Cref{app:topic_judge}.

\paragraph{Results.}
We perform 100 comparisons for each model pair (2800 comparisons per dataset) using GPT-4.1 \citep{openai2025gpt4-1} with temperature 0, prompting the LLM to choose the set of topics that best captures the general meaning of the document.

\Cref{tab:topic_judge_standard,tab:topic_judge_hard} shows that an MTM achieves the highest Elo score in 14 of 16 dataset--representation rows, and MTMs outperform their word-based counterparts in 42 of 48 pairwise comparisons. Of all 48 comparisons, 71\% show an MTM advantage of at least 50 points (${\sim}57\%$ win probability), 50\% exceed 100 points (${\sim}64\%$), and 21\% exceed 200 points (${\sim}76\%$). The main exception is Bills with top features, where all three word-based models outscore their MTM counterparts; we attribute this to the judge finding MTM's SAE feature descriptions less specifically relevant than the precise keywords learned by word-based models (\Cref{sec:methods:featurization}). Switching to the summarization representation recovers mLDA and mETM on Bills, though mBERTopic remains lower. The advantages are largest on the challenging datasets---GoEmotions, PoemSum, and WritingPrompts---where MTMs win by an average of 195 Elo points, suggesting that SAE features capture semantic nuances beyond word co-occurrence patterns. %

To validate the LLM judge, we conducted a human evaluation study on Wiki and PoemSum. We recruited 68 participants from an adult population with at least a bachelor's-level English education. Each participant was shown a document alongside topic representations from multiple models and asked to judge which topics best capture the document's meaning. For statistical efficiency, each participant ranked 5 randomly sampled models per document across 10 tasks; %
we then decomposed rankings into pairwise comparisons and fit a Bradley-Terry model to obtain Elo ratings. Further details are provided in \Cref{app:human_evaluation}.

\Cref{tab:human_topic_judge} reports the resulting human Elo scores. MTMs achieve the highest scores in three of four conditions; the exception is PoemSum with summaries, where BERTopic ranks first but is not significantly different from MTM (w/ ETM). To assess agreement between human and LLM evaluators, we plot their Elo scores against each other in \Cref{fig:human_llm_elo} and compute Spearman's $\rho$ for each condition. The points cluster closely around the diagonal, and the scores are strongly correlated ($\rho$ between 0.76 and 0.95), with agreement holding for both MTMs and word-based models. This confirms that the LLM judge is a reliable proxy for human preferences.

\subsection{Standard Evaluation Metrics}
\label{sec:empirical_studies:coherence_td}

\begin{figure*}[t]
  \centering
  \includegraphics[width=\linewidth]{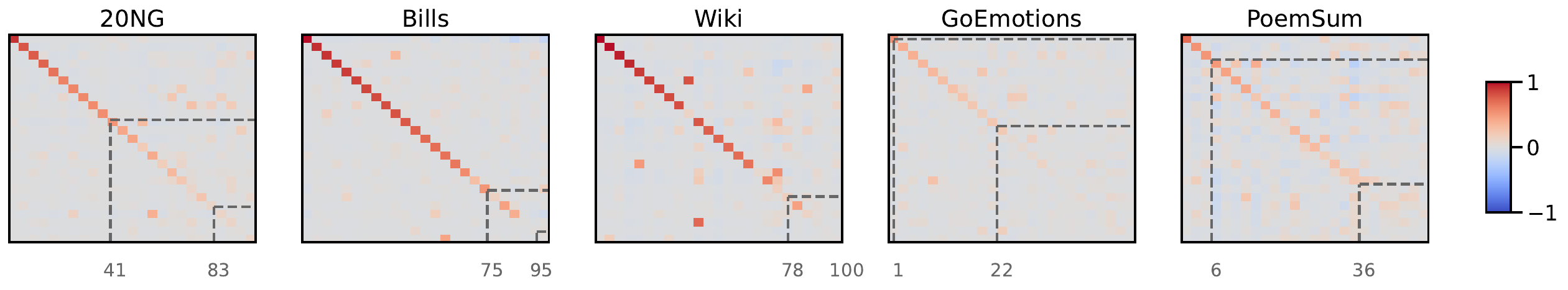}
  \caption{Heatmap representations of how similar mLDA topics (columns) and LDA topics (rows) are in terms of their proportions across documents; topics are aligned greedily. The two models learn similar topics on Bills and Wiki,  but mLDA finds new topics on 20NG, GoEmotions, and PoemSum. Dashed gray boxes show submatrices: within, the two models' topics have medium ($\leq 0.5$) or low ($\leq 0.2$) correlations.}
  \label{fig:topic_correlations}
\end{figure*}

While topic judge is a holistic measure of model performance, we also report two complementary metrics: \textit{coherence}—the semantic relatedness of a topic's top features—and \textit{topic diversity}—the distinctiveness of topics. 

We measure coherence in two ways. First, we measure the average rating assigned to each topic on a 1–3 scale based on how semantically related its top features are (R) \citep{newman2010automatic}. Second, we measure how accurately an evaluator can identify an ``intruder'' feature from another topic when it's mixed with the target topic's features (I) \citep{chang2009reading}. Following \citet{stammbach-etal-2023}, we use GPT-4.1 as the rater and evaluator in both tasks (see \Cref{app:all_prompts:coherence} for the prompts).  For ratings, we present the top 10 features per topic and report the average rating across all topics. For intrusion, we perform 25 trials with 5 true features and 1 intruder per trial and report the average accuracy. For both, we use zero temperature to sample.

To measure topic diversity (TD), we use the metric by \citet{dieng2020topic} described in \Cref{app:topic-diversity}.

Results in \Cref{tab:coherence_td_standard} show that MTMs achieve 2.5--2.9 coherence on benchmark datasets, although word‐based models outperform MTMs in some cases. On the challenging datasets, MTMs perform well, indicating that SAE directions remain interpretable even for very short texts (GoEmotions) or abstract themes (PoemSum, WritingPrompts). In topic diversity, mLDA and mETM are comparable to word-based models (\Cref{tab:coherence_td_short}). However, mBERTopic learns some topics with substantial overlap on PoemSum and WritingPrompts (0.38 and 0.36 diversity) and has a lower coherence score of 2.28 on PoemSum. Further work is needed to improve mBERTopic's robustness across datasets.

Finally, we report topical alignment (\Cref{tab:alignment}) and model stability (\Cref{tab:stability}) in the Appendix. MTMs achieve comparable alignment compared to word-based models on benchmark datasets, and learn topics that are as or more stable than their word-based counterparts under the stability metric of \citet{hoyle-etal-2022-neural}.

\begin{table*}[h]
\centering
\footnotesize
\begin{tabular*}{\textwidth}{@{\extracolsep{\fill}}p{1.2cm}p{\dimexpr\textwidth-1.2cm\relax}@{}}
\toprule
\multicolumn{2}{@{}p{14cm}@{}}{\textbf{Topic Summary:} "Achievements, statistics, and aspirations in college and professional sports, focusing on tournaments, records, player performance, and team accomplishments."} \\
\midrule
$\lambda=10$ & "A text about the history of a place, its people and their customs.\textbackslash n\textbackslash nThe book is written in an easy language, with many illustrations to make it more interesting for children. It also contains some words in the local dialect [...]"\\
\midrule
$\lambda=20$ & "A text about the 2019 novel coronavirus (COVID-19) outbreak in China has been circulating on WhatsApp. The message claims that a new strain of the virus, which is more deadly than SARS and MERS, has emerged from [...]"\\
\midrule
$\lambda=30$ & "A text about the 2018 NBA draft has been circulating on social media, and it's a doozy.\textbackslash n\textbackslash nThe text is from an anonymous source who claims to have inside information that says Zion Williamson will be drafted by [...]"\\
\midrule
$\lambda=40$ & "A text about the 2014-15 NBA All Star Game was announced on Thursday.\textbackslash n\textbackslash nThe game will be held at Madison Square Garden in New York City, and it will feature a team of all stars from both leagues. The game is scheduled." \\
\midrule
$\lambda=50$ & "A text about the best-selling player in college basketball, a 10th place finish and an All-American selection.\textbackslash n\textbackslash nThose are just some of the accomplishments that have been achieved by Baylor's senior forward John Wall."\\
\midrule
\end{tabular*}
\caption{Examples of generated text at various steering strengths ($\lambda$) for a sports-related topic found by mLDA on the Wiki dataset. Higher $\lambda$ values result in increasingly topic-focused content.}
\label{tab:steered_text_mlda_main}
\end{table*}

\subsection{Topic Novelty Analysis}
\label{sec:empirical_studies:qualitative}
Having established that MTMs are competitive on standard metrics and preferred by topic judge, we now investigate whether they discover new topics or instead redescribe topics that word-based models already find. To address this, we analyze how similarly topics from different models weigh documents---if two topics assign similar weights across documents, they likely capture the same concepts regardless of description.

We formalize this idea by computing correlations between document-topic distributions. Consider trained models $m$ and $m'$ with document-topic matrices $\thetaa^m, \thetaa^{m'} \in \mathbb{R}_+^{D \times K}$, where row $d$ contains document $d$'s topic weights and column $\thetaa^m_{:, i}$ is topic $i$'s prevalence across documents. We compute the cross-correlation matrix $\mathbf{C} \in \mathbb{R}^{K \times K}$, where each entry is $C_{i,j} = \text{corr}(\thetaa^m_{:, i}, \thetaa^{m'}_{:, j})$.

\Cref{fig:topic_correlations} shows these correlations after greedy alignment: we iteratively select the unpaired column with the strongest correlation in the entire matrix and pair it from the remaining rows, placing matches consecutively along the diagonal (see \Cref{app:alignment} for details). Dashed boxes mark regions where correlations fall below 0.5 or 0.2, highlighting where MTMs discover different topics.

On Bills and Wiki, over 70\% of mLDA topics correlate at least 0.5 with LDA topics, suggesting that the topics represent similar themes. On GoEmotions and PoemSum, nearly all topics fall below 0.5 correlation (many below 0.2), indicating they are new. 20NG lies between these extremes---we think this is due to MTMs capturing stylistic qualities like argumentation that word co-occurrence misses. ETM and BERTopic pairs show similar patterns. The three additional datasets (\Cref{fig:topic_correlations_new}) also vary: AGNews shows moderate overlap, while Yelp and WritingPrompts topics are largely novel.
We provide examples of MTM topics, including novel topics, in \Cref{fig:new-topics}.

\subsection{Steering Evaluation}
\label{sec:empirical_studies:steering}
MTMs enable text generation via steering vectors formed from discovered topics (\Cref{sec:method:steering}). 
To evaluate this capability, we conduct three experiments. We measure two criteria for steering: \textit{topic relevance}---steering increases the expression of the target topic in text; and \textit{fluency}---steering preserves the coherence and naturalness of the generated text.

Throughout, we use $p(\cdot)$ and $p(\cdot; \lambda\mathbf{s}_k)$ for unsteered and steered LLM probabilities, respectively (steering vector $\mathbf{s}_k$, magnitude $\lambda$), $\mathbf{x} = (x_1, \ldots, x_{N_{\text{gen}}})$ for sampled token sequences, and $\mathbf{x}_d = (x_{d,1}, \ldots, x_{d,N_{\text{tok}}})$ for document $d$ sequences. We describe each experiment below, with additional details in \Cref{app:steering_details}.

\paragraph{Topic Relevance Win Rate.} We first verify whether MTM steering vectors guide LLM-generated text toward exhibiting specific topics. We take a ``best-of-$2L$'' approach: for each topic $k$, we sample $L$ steered texts with different steering strengths and $L$ unsteered texts using the same prompt. An LLM judge then selects the text most representative of the topic according to its summarized description $\texttt{t}_k$. We record a win if any steered text is chosen. %

We repeat this procedure $R$ times for each topic. We define the topic relevance win rate (TWR) as the fraction of comparisons where the judge selects one of the $L$ steered samples, averaged across the $KR$ trials. When computed for a single topic, a TWR greater than 0.5 indicates that steering biased text generation toward the specified topic.

\paragraph{Topic Likelihood Difference.}
We next assess if steering vectors capture topic semantics by comparing their effect on the likelihood of different documents from the training corpus. Intuitively, if $\mathbf{s}_k$ represents topic $k$, increasing the steering strength $\lambda$ should increase the likelihood of documents about topic $k$ more than other documents from the corpus.

Formally, let $\mathcal{D}_k$ contain documents highly associated with topic $k$ and $\mathcal{D}_{-k}$ be an equally sized random sample from the set of documents highly associated with another topic (see \Cref{app:steering:topic_likelihood_diff}). We measure the relative log likelihood difference when steering as
\begin{equation}
\Delta\ell_k(\lambda) = \frac{1}{S}\sum_{i=1}^S \log \frac{p(\mathbf{x}_{d^+_i}; \lambda\mathbf{s}_k)}{p(\mathbf{x}_{d^-_i}; \lambda\mathbf{s}_k)},
\end{equation}
where $d^+_i \in \mathcal{D}_k$ and $d^-_i \in \mathcal{D}_{-k}$.
A positive value for $\Delta\ell_k(\lambda)$ when $\lambda \gg 0$ indicates effective topical steering, while a negative value when $\lambda = 0$ indicates effective topic ablation.
We let $\Delta \ell(\lambda)$ without subscript denote the average over all topics. 

We also use this metric to evaluate partial steering vectors $\mathbf{s}_k^{(n)}$, constructed from only the top $n$ features in the topic, to verify the advantage of using the full steering vector.

\paragraph{Perplexity.} Finally, we evaluate whether steering maintains natural-sounding generations. We compute the perplexity (PPL) of the sampled steered generations $\x' \sim p(\x' ; \lambda \s_k)$  under the original model $p(\x)$. 
We report the perplexity for 10 generations per topic under five values of $\lambda$, and the perplexity of unsteered generations as a baseline. Values close to the baseline indicate that fluency is comparable to that of the unsteered texts.

\begin{figure*}[t]
  \centering
  \includegraphics[width=\textwidth]{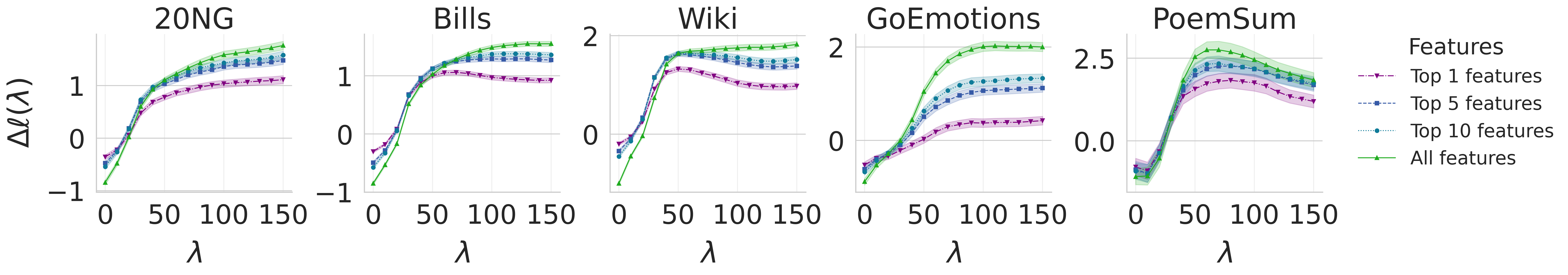}
  \caption{Average document log likelihood differences $\Delta\ell(\lambda)$ between on-topic and off-topic documents for mLDA. Lines show $\Delta \ell(\lambda)$ with steering vectors $\s_k^{(n)}$ constructed using $n$ = 1, 5, 10, or all SAE features. Negative $\Delta\ell$ at $\lambda=0$ shows that the topics are ablated; positive $\Delta\ell$ at higher $\lambda$ demonstrates steering toward target topics, with better steering when using all features.}
  \label{fig:log-prob-graphs}
\end{figure*}

\begin{table}[t]
  \centering
  \resizebox{\columnwidth}{!}{%
  \footnotesize
  \begin{tabular}{@{}lcccccccc@{}}
    \toprule
      & 20NG & Bills & Wiki & Yelp & AGN & GoEmo & Poem & WP \\
    \midrule
    TWR (\%)
      & 88.7
      & 98.9
      & 95.1
      & 92.6
      & 96.6
      & 84.4
      & 85.8
      & 87.7 \\
    \midrule
    $\text{PPL}_\text{control}$
      & 6.23
      & 6.24
      & 6.24
      & 6.23
      & 6.24
      & 6.25
      & 6.23
      & 6.22 \\
    $\text{PPL}_{\lambda=10}$
      & 6.36
      & 6.27
      & 6.43
      & 6.40
      & 6.34
      & 6.31
      & 6.46
      & 6.33 \\
    $\text{PPL}_{\lambda=20}$
      & 6.93
      & 6.83
      & 7.04
      & 7.14
      & 6.72
      & 6.61
      & 7.19
      & 7.00 \\
    $\text{PPL}_{\lambda=30}$
      & 7.97
      & 8.04
      & 7.72
      & 8.62
      & 7.49
      & 7.83
      & 8.89
      & 8.24 \\
    $\text{PPL}_{\lambda=40}$
      & 9.82
      & 9.42
      & 9.23
      & 11.14
      & 8.61
      & 10.68
      & 11.55
      & 10.82 \\
    $\text{PPL}_{\lambda=50}$
      & 14.52
      & 12.49
      & 13.03
      & 17.11
      & 11.02
      & 20.13
      & 16.86
      & 16.95 \\
    \bottomrule
  \end{tabular}%
  }%
\caption{Steering metrics for mLDA across all 8 datasets. TWR: fraction of times steered text better matches target topic. PPL: perplexity of steered generations under the original model at various steering strengths $\lambda$.}
  \label{tab:winrates}
\end{table}

\subsubsection{Results}
\Cref{tab:steered_text_mlda_main} shows representative examples of text steered toward a sports-related topic. This topic was discovered by mLDA trained on the Wiki dataset, and the table illustrates outputs generated using a range of steering strengths ($\lambda$). At lower values of $\lambda$, generated texts remain generic and off-topic, while at higher values, the content aligns with the sports theme. Additional generated texts for other models, datasets, and steering strengths are provided in \Cref{tab:steered_text_examples}.

Quantitatively, steering improves topic relevance. As shown in \Cref{tab:winrates}, the topic relevance win rate (TWR) exceeds 84\% across all datasets, reaching 99\% on Bills. This indicates that steering guides text generation toward intended topics. \Cref{tab:twr-by-magnitude} shows that higher steering strengths win more often: $\lambda\in\{30, 40, 50\}$ accounts for 80\% of mLDA wins, 77\% of mETM wins, and 86\% of mBERTopic wins. \Cref{fig:log-prob-graphs} further illustrates this effect by showing how relative document log likelihood changes as steering strength increases. At baseline ($\lambda=0$), ablation reduces log likelihood for on-topic documents. Moreover, positive steering ($\lambda>0$) increases the likelihood of on-topic documents relative to off-topic ones. Steering vectors that use all topic features produce the largest differences, indicating that the full feature set effectively characterizes topic expression. \Cref{fig:mlda-log-probs-right-third} shows log likelihood plots for mLDA on Yelp, AGNews, and WritingPrompts; \Cref{fig:combined-log-probs} shows the corresponding mETM and mBERTopic plots. %

Finally, the perplexities of steered outputs (\Cref{tab:winrates}) remain close to baseline values, showing that steering preserves the fluency of generated text. The steering metrics for mBERTopic and mETM are reported in \Cref{tab:mbertopic-twr-ppl-results,tab:all-twr-ppl-results}.

%% file: sections/6-discussion.tex
\section{Limitations}
\label{sec:limitations}
\raggedbottom

MTMs trade the simplicity of word-based topic models for richer semantic representations, but this tradeoff introduces two practical constraints.

The first constraint is a dependency on SAEs and their ecosystem. Pretrained SAEs are currently available for only a handful of LLMs, and topic quality is sensitive to the choice of SAE layer, the dataset used to train the SAE, and the accuracy of automatically generated feature descriptions. Hence, if MTMs use an SAE that is not sufficiently relevant or specific for the dataset at hand, or if the labels for the features are inaccurate, the resulting topic quality may be low. We believe training or finetuning SAEs directly on target corpora, as in \citet{movva2025saehypothesis}, is a promising direction to mitigate these issues, as features learned from the data of interest are more likely to be relevant and their auto-generated labels more accurate.

The second constraint is a heavier preprocessing pipeline. Although the underlying algorithms (LDA, ETM, BERTopic) scale comparably to their word-based counterparts once given a feature matrix, the pipeline adds upfront featurization through an LLM and SAE, and feature filtering via another LLM (GPT-4o mini in our experiments, at a total cost of \$0.33 for the eight datasets). Featurization requires a GPU with enough memory to fit the LLM and SAE. Depending on the dataset, this can represent a significant additional cost compared to simpler alternatives; \Cref{tab:featurization_time} reports per-dataset throughput to provide an idea of additional runtime. While these additional demands are not prohibitive and need not preclude practitioners from using MTMs, they should be weighed when deciding whether richer topic descriptions, topic steering, and more abstract topics justify the overhead.

\section{Conclusion}
\label{sec:conclusion}
We introduced Mechanistic Topic Models (MTMs), a family of topic models that operate on SAE activation patterns rather than word counts or raw text embeddings.
Our empirical evaluation shows MTMs are comparable with baselines on standard benchmarks, with significant advantages on abstract and short-text datasets as measured by topic judge. MTMs enable controlled text generation, allowing researchers to create new texts with specific topic compositions.

MTMs suggest how, despite some recent negative results, interpretability tools like SAEs can be successfully repurposed for downstream tasks.

%% file: sections/appendices/mtm_implementation_details.tex
\section{MTM Implementation Details}
\label{app:mtm_implementation_details}

\Cref{app:preprocessing_filtering} and \Cref{app:posttraining_refinement} detail the quality control measures used to address challenges in using SAE features in MTMs.

\subsection{Preprocessing: Feature Filtering}
\label{app:preprocessing_filtering}

Before training MTMs, we filter out SAE features that are likely spurious or irrelevant for topic modeling. We remove features in the following categories:

\begin{itemize}
    \item Features with textual descriptions about programming, math, grammar, text formatting, or stop words.
    \item Features that activate in more than 1\% of the SAE training data.
    \item Features that appear in more than 90\% of corpus documents (analogous to removing high-frequency words in traditional topic modeling).
\end{itemize}
This filtering step is crucial.  Without it, the resulting models do poorly in our benchmarks. 
To remove the features with irrelevant textual descriptions, we use the OpenAI GPT-4o mini model \citep[version 2024‑07‑18]{openai2024gpt4omini}.

\subsection{Post-training: Topic Refinement}
\label{app:posttraining_refinement}

Even after preprocessing, SAE feature descriptions can occasionally be mislabeled. These mislabelings often become apparent when examining the full textual representation $\texttt{t}_k$ of a topic (see \Cref{sec:methods:interpreting_topics}). 
An example of this type of mislabeling is provided in \Cref{fig:example_topics:mislabeled}.

To address this, we apply an automated post-training refinement step. This step is optional and can easily be done in practice by inspection.
\begin{enumerate}
    \item For each topic $k$, retrieve the top $n+m$ features by weight (where $m$ is small, e.g., 2).
    \item Prompt an LLM to identify and remove up to $m$ features that appear irrelevant or mislabeled relative to the other features.
    \item Retain the resulting top $n$ features as the final topic representation.
\end{enumerate}

Automated refinement is cost-effective since the number of topics and features per topic is typically small. The specific prompt used for this task is provided in \Cref{app:all_prompts:topic_refinement}. We set $n=10$ and $m=2$ in our experiments.

To ensure a fair comparison, and avoid biasing the coherence metrics in favor of MTMs, we apply the exact same post-training refinement to all models in our experiments, including word-based baselines (see Section~\ref{sec:empirical_studies}).

%% file: sections/appendices/datasets.tex
\section{Datasets}
\label{app:datasets}
\begin{table*}[t]
  \centering
  \small
  \begin{tabular}{@{}l r rrr rrr@{}}
    \toprule
    & & \multicolumn{3}{c}{Word-based} & \multicolumn{3}{c}{Mechanistic} \\
    \cmidrule(lr){3-5} \cmidrule(lr){6-8}
    Corpus & \#Docs
           & \#Words     & $|V|$  & Avg.\ Len.
           & \#Tokens    & $|V|$  & Avg.\ Len. \\
    \midrule
    20NG           & 10,496 &  780,825   & 15k &  74.4 & 3,023,135 & 8,750 & 288.0 \\
    Bills          & 32,659 & 3,442,488  & 15k & 105.4 & 7,729,695 & 8,750 & 236.7 \\
    Wiki           & 14,290 & 14,037,490 & 15k & 982.3 & 38,065,712 & 8,746 & 2663.8 \\
    GoEmotions     & 19,264 &  131,293   &  5k &   6.8 &   767,694 & 8,750 &  39.9 \\
    PoemSum        &  2,398 &  184,695   &  5k &  77.0 &   707,794 & 8,743 & 295.2 \\
    AGNews         & 10,000 &  162,414   &  5k &  16.2 &   528,485 & 8,750 &  52.8 \\
    Yelp Polarity  & 10,000 &  465,280   & 15k &  46.5 & 1,698,692 & 8,749 & 169.9 \\
    WritingPrompts & 10,000 & 2,063,889  & 15k & 206.4 & 6,924,757 & 8,749 & 692.5 \\
    \bottomrule
  \end{tabular}
  \caption{Training corpus statistics after preprocessing for the word-based and mechanistic topic models.}
  \label{tab:app:corpus_stats_combined}
\end{table*}

The dataset statistics are in \Cref{tab:app:corpus_stats_combined}. For the word-based models, the documents are preprocessed into word counts using the \texttt{soup-nuts} package \citep{hoyle2021automated} with the following settings: words are lowercased and named entities are automatically detected and merged together using spaCy \citep{honnibal2020spacy} (e.g., ``New York'' becomes ``New\_York''); words must match the regex \verb|^[\w-]*[a-zA-Z][\w-]*$|, contain at least 2 characters, and appear in less than 90\% of documents. The vocabulary size is set to 5k for the smaller datasets (AGNews, GoEmotions, PoemSum) and 15k for the remaining datasets. After preprocessing, documents with less than 5 words are removed from the training corpus. We randomly subset to 10,000 training documents for AGNews, Yelp Polarity, and WritingPrompts.

\paragraph{Labels.}
Below is the label information for each dataset, along with some examples of labels.

\textbf{20NG} (20 categories): \textit{talk.politics.guns, comp.graphics, misc.forsale}.

\textbf{Bills} (114 subtopics): \textit{drug coverage and cost, water resources, insurance, postal service}.

\textbf{Wiki} (279 subcategories): \textit{architecture buildings, 1990--1999 songs, fungi, warships of Russia}.

\textbf{GoEmotions} (28 annotated emotions): \textit{anger, disappointment, optimism, neutral}.

%% file: sections/appendices/baseline_implementations.tex
\section{Baseline Implementations}
\label{app:baseline_implementations}
We implement the baselines as follows. For LDA, we default to using the \texttt{Mallet} package \citep{mallet}, which implements Gibbs sampling for inference. For the Wiki dataset, due to the large number of words/tokens, we instead use the LDA \texttt{scikit-learn} implementation \citep{scikit-learn} with coordinate ascent variational inference. %
For the four neural models, we use the authors' provided code when it is available---we use the \texttt{bertopic} and \texttt{fastopic} PyPI packages, and for ETM and D-VAE, we use the authors' provided model code and re-implement the model training in our codebase using PyTorch Lightning.

For the baseline models, we use the default settings from either the paper or codebase (if unspecified in the paper), unless detailed as follows.

For LDA, we use the default settings in the \texttt{Mallet} package.

For ETM, we specifically use the Labeled ETM variant (i.e., we train skip-gram embeddings on the dataset and initialize the word embeddings with them, which are frozen during training), as we found that it outperformed regular ETM.

For BERTopic and FASTopic, we use the all-mpnet-base-v2 embedding model. For BERTopic, we set the top $n$ words for c-TF-IDF to 25.

We allow ETM and D-VAE to use a separate validation set for early stopping. For mechanistic ETM, we instead form a validation set from 10\% of the training set.

%% file: sections/appendices/bayesian_optimization.tex
\section{Bayesian Optimization}
\label{app:bayesian_optimization}

We run Bayesian optimization for 25 iterations for each model-dataset pair. The first subsection lists the hyperparameter search space for each model. The next subsections provide mathematical formulations of the topic quality metrics used in the optimization objective. Following \citet{dieng2020topic}, we set the optimization objective to be the product of NPMI coherence and topic diversity.

\subsection{Hyperparameter Ranges}
Here, $\llbracket a, b \rrbracket$ denotes the set of integers from $a$ to $b$.

\par\vspace{1\baselineskip}

\sisetup{
  list-separator = {,},       %
  list-pair-separator = {,},  %
  list-final-separator = {,}  %
}

D-VAE
\begin{description}[nosep,font=\normalfont]
  \item[] $\texttt{topic density} \in [0.01, 5.0]$ (log-uniform)
  \item[] $\texttt{learning rate} \in [10^{-3}, 10^{-1}]$ (log-uniform)
  \item[] $\texttt{n KL divergence warmup epochs} \in \llbracket 100, 200 \rrbracket$
  \item[] $\texttt{use RSVI} \in \{\texttt{true}, \texttt{false}\}$
\end{description}

FASTopic
\begin{description}[nosep,font=\normalfont]
    \item[] $\texttt{DT alpha} \in \llbracket 1, 25 \rrbracket$
    \item[] $\texttt{TW alpha} \in \llbracket 1, 25 \rrbracket$
    \item[] $\texttt{n epochs} \in \llbracket 100, 400 \rrbracket$
    \item[] \mbox{$\texttt{Sinkhorn threshold} \in [10^{-7}, 0.05]$} (log-uniform)
\end{description}

LDA \& mLDA
\begin{description}[nosep,font=\normalfont]
  \item[] $\texttt{topic density} \in [0.01, 5.0]$ (log-uniform)
  \item[] $\texttt{word/feature density} \in [0.01, 0.1]$ (log-uniform)
\end{description}

ETM
\begin{description}[nosep,font=\normalfont]
  \item[] $\texttt{learning rate} \in [10^{-4}, 10^{-2}]$ (log-uniform)
  \item[] $\texttt{weight decay} \in [10^{-7}, 10^{-5}]$ (log-uniform)
  \item[] $\texttt{use doc. completion validation} \in \{\texttt{true}, \texttt{false}\} $
\end{description}

mETM
\begin{description}[nosep,font=\normalfont]
  \item[] $\texttt{learning rate} \in [10^{-3}, 10^{-2}]$ (log-uniform)
  \item[] $\texttt{weight decay} \in [10^{-7}, 10^{-5}]$ (log-uniform)
  \item[] $\texttt{dropout} \in [0, 0.1]$
  \item[] $\texttt{decoder dropout} \in [0, 0.1]$
  \item[] $\texttt{scheduler type} \in \{\texttt{none}, \texttt{cosine}\} $
\end{description}

BERTopic
\begin{description}[nosep,font=\normalfont]
    \item[] $\texttt{n UMAP neighbors} \in \llbracket 5, 50 \rrbracket$
    \item[] $\texttt{n UMAP components} \in \llbracket 5, 50 \rrbracket$
    \item[] $\texttt{min topic size} \in \llbracket 5, 15 \rrbracket$
\end{description}

mBERTopic
\begin{description}[nosep,font=\normalfont]
    \item[] $\texttt{n UMAP neighbors} \in \llbracket 5, 50 \rrbracket$
    \item[] $\texttt{n UMAP components} \in \llbracket 5, 50 \rrbracket$
    \item[] $\texttt{min topic size} \in \llbracket 5, 15 \rrbracket$
    \item[] $\texttt{use SAE count embeddings} \in \{\texttt{true}, \texttt{false}\}$
    \item[] \mbox{$\texttt{use Sentence Transformer embs.} \in \{\texttt{true}, \texttt{false}\}$}\footnote{This option was always chosen to be false.}
\end{description}

\subsection{NPMI Coherence}

NPMI (Normalized Pointwise Mutual Information) coherence, or $C_{\text{NPMI}}$, measures the semantic relatedness of the top words within each topic based on their co-occurrence patterns in a reference corpus. For a given topic, NPMI coherence is defined as

\begin{equation} C_{\mathrm{NPMI}} = \frac{1}{\binom{10}{2}} \sum_{j=2}^{10}\sum_{i=1}^{j-1} \frac{ \log \frac{p(w_i,w_j)}{p(w_i)p(w_j)} }{ -\log p(w_i,w_j) },
\end{equation}

where $\{w_i\}_{i=1}^{10}$ are the top 10 words in a topic, and $p(w_i, w_j)$ is estimated using word co-occurrences in a sliding context window across documents \citep{lau2014machine}. In our experiments, we set the reference corpus to be the training corpus, and the entire document is used as the context window. The probabilities $p(w_i)$ and $p(w_j)$ represent the individual word frequencies in the corpus.

\subsection{Topic Diversity}
\label{app:topic-diversity}

Topic diversity measures how distinct topics are from each other by computing the fraction of unique features among the top features across all topics. Formally, topic diversity is calculated as

\[
\mathrm{TD}
=
\frac{\left|\bigcup_{k=1}^{K}\mathrm{TopFeatures}_k^{(n)}\right|}{Kn}.
\]

Let $\mathrm{TopFeatures}_k^{(n)}$ denote the $n$ highest-weight features of topic $k$, where features are words for word-based models and SAE features for mechanistic models. Following \citet{dieng2020topic}, we use $n=25$.

\subsection{Limitations for Cross-Model Comparison}

While topic diversity is roughly comparable across mechanistic and word-based topic models, NPMI coherence is not directly comparable between these model types due to their differing vocabulary distributions. Mechanistic models operate on SAE feature spaces, while word-based models use traditional word vocabularies, making direct coherence comparisons problematic.

%% file: sections/appendices/steering_experiment_details.tex
\section{Steering Experiment Details}
\label{app:steering_details}

\subsection{Steering Intervention}
For all steering, we perform the topic ablation and subsequent topic addition interventions across all layers and all token positions using a pre-hook on the forward activations. 

\subsection{Topic Relevance Win Rate}
\label{app:steering:twr_llm_judge}
We run $R = 10$ trials per topic with temperature set to 0.3 for both the steered and unsteered models. We generate a maximum of 50 tokens. 
We use GPT-4o mini \citep{openai2024gpt4omini} as the LLM judge, and the prompt is provided in \Cref{app:steering:twr_llm_prompt}.

\subsection{Topic Likelihood Difference}
\label{app:steering:topic_likelihood_diff}
We select the most relevant documents for each topic using a threshold-based approach. For threshold $\tau = 0.5$, we define $\widetilde{\mathcal{D}}_k = \{d : \thetaa_{d,k} \geq \tau\}$ and set
\[
\mathcal{D}_k =
\begin{cases}
\widetilde{\mathcal{D}}_k, & 3 \le |\widetilde{\mathcal{D}}_k| \le 10\\
\mathrm{Top}_{10}(\widetilde{\mathcal{D}}_k), & |\widetilde{\mathcal{D}}_k| > 10\\
\mathrm{Top}_{3}(\{1,\dots,\mathcal{D}\}). & |\widetilde{\mathcal{D}}_k| < 3
\end{cases}
\]
This ensures 3--10 documents per topic. We use all threshold-exceeding documents when feasible, capping at 10 for topics with many relevant documents. When fewer than 3 documents meet the threshold, we select the top 3 from the corpus.

%% file: sections/appendices/all_prompts.tex
\section{Prompts}
\label{app:all_prompts}
We provide the prompt templates used for post-training topic refinement (\Cref{app:posttraining_refinement}), topic summarization (\Cref{sec:methods:interpreting_topics}), and the LLM-based evaluations in \Cref{sec:empirical_studies}. For the topic judge prompt and examples, see \Cref{app:topic_judge}. 

The prompted LLM is GPT-4.1 \citep[version 2025‑04‑14]{openai2025gpt4-1}, except for topic relevance win rate (see \Cref{app:steering:twr_llm_judge}).

\subsection{Topic Refinement}
\label{app:all_prompts:topic_refinement}

\extrasmall{
\textbf{System prompt.} The goal of this task is to evaluate a list of features produced by an automatic method. We call this list a "topic". Given a topic, you'll be answering the question: "Which [\textit{word | feature}](s) don't belong in this list?" For each topic, choose the [\textit{word | feature}](s) whose meaning does not match with what the list seems to be about.

Here is an example: [example]

Here is another example: [example]

Reply with a brief reasoning for your choice, and up to two numbers corresponding to the [\textit{word | feature}]s that don't belong (or -1 if there are less than two).

Important: Prioritize identifying [\textit{word | feature}](s) that are oddly specific and/or clearly out of place. Use your world knowledge in considering whether a [\textit{word | feature}] belongs. If you are not sure, do not choose the [\textit{word | feature}].
}

\noindent\extrasmall{
\textbf{User prompt.} Topic: [topic]
}

\subsection{Topic Summarization}
\label{app:all_prompts:topic_summarization}
\extrasmall{
\textbf{System prompt.} 
You are a helpful assistant specializing in topic summarization. You will be given a list of either topic keywords (some of them may be several words concatenated with "\_") or descriptions of text generated by an automated method. Your task is to summarize these keywords or descriptions into no more than 1 sentence describing the topic's central theme.

Provide a concise and informative summary that captures the topic's essence. Here are some specific guidelines:

1. Do not use a full sentence or a complete thought.

2. Use your world knowledge to help you decide what the topic is about.

3. The summary should be general, capturing the commonalities of the items as a single main theme. In particular, do not rely on lists in your response, or include specifics that only pertain to a few items in the topic.

4. If unsure, err on the side of being more general rather than too specific in your summary.
}

\noindent\extrasmall{
\textbf{User prompt.} Topic: [topic]
}

\subsection{Ratings and Intrusion}
\label{app:all_prompts:coherence}

\extrasmall{
\textbf{System prompt.}
The goal of this task is to evaluate a list of [\textit{word | feature}]s produced by an automatic method. We call this list a "topic". Given a topic, you will determine how related its [\textit{word | feature}]s are on a 3-point scale. The rating options are: (1) Not Very Related, (2) Somewhat Related, (3) Very Related. A helpful question to ask yourself is: "What is this group of [\textit{word | feature}]s about?" If you can answer easily, then the [\textit{word | feature}]s are probably related. Use your world knowledge and the context provided by the other [\textit{word | feature}]s to help determine your rating. Here is some guidance and examples on how to apply these ratings.

Very Related - Most of the [\textit{word | feature}]s are clearly related to each other, and it would be easy to describe how they are related.

Example 1: [example]

Example 2: [example]

Somewhat Related - The [\textit{word | feature}]s are loosely related to each other, but there may be a few that are ambiguous, generic, or unrelated.

Example 1: [example]

Example 2: [example]

Not Very Related - The [\textit{word | feature}]s do not share any obvious relationship to each other. It would be difficult to describe how the [\textit{word | feature}]s are related to each other.

Example: [example]

Reply with a brief reasoning for your choice and a single number, indicating the overall relatedness of the [\textit{word | feature}]s in that topic.
}

\noindent\extrasmall{
\textbf{User prompt.} Topic: [topic]
}

\par\vspace{1\baselineskip}

\noindent\extrasmall{
\textbf{System prompt.}
The goal of this task is to evaluate a list of [\textit{word | feature}]s produced by an automatic method. We call this list a "topic". Given a topic, you'll be answering the question: "Which [\textit{word | feature}] doesn't belong in this list?" For each topic, choose the [\textit{word | feature}] with the meaning or usage that is most different from the others. If you feel that multiple [\textit{word | feature}]s do not belong, choose the one that you feel is most out of place.

Here are some examples: [example]

Here is another, harder example: [example]

You might be given multiple topics. For each topic, reply with a brief reasoning for your choice and the number of the [\textit{word | feature}] that doesn't belong.
}

\noindent\extrasmall{
\textbf{User prompt.}
Topic: [topic]
}

\subsection{Topic Relevance Win Rate (TWR)}
\label{app:steering:twr_llm_prompt}
\extrasmall{
\textbf{System prompt.} You are an expert evaluator of text relevance to topics. You will be given a topic summary and a list of text samples. Your task is to determine which text sample is most relevant to the given topic.}  

\noindent\extrasmall{\textbf{User prompt.} Topic summary: [topic summary]

Text samples:
[texts]

Which text sample (by index number) is most relevant to the topic? Provide the index (starting from 0) and a brief explanation.
}

\normalsize

%% file: sections/appendices/topic_judge.tex
\section{Topic Judge}
\label{app:topic_judge}

In our experiments, we set $q=2$, $p=0.75$, and $T=100$. For TopFeatures, we always take the top 10 words for baselines, and take either the top 10 (for documents with 1 topic) or top 5 (for documents with 2 topics) features for MTMs.

We provide the prompt and example inputs and responses for topic judge below.

\subsection{Prompt}

\extrasmall{
\textbf{System prompt.} In this task, you will be presented with a document, a criteria, and two sets of "topics". [\textit{A given topic is a list of either single words (or occasionally, instead of a single word, several words concatenated with "\_") or descriptions of text about 5-15 words long. | A given topic will be shown as a summary description no more than 1 sentence long.}] Each set of topics includes 1-2 topics total. The task is to choose which of the two sets of topics is better suited to the document based on the provided criteria. Reply with "A" if the first set of topics is better or "B" if the second set of topics is better. If you think that the two sets of topics are equally good, please reply with "tie". Only use "tie" if the two sets of topics are very similar and you cannot choose one over the other. Before making your choice, provide a brief reasoning for your decision.
}

\noindent\extrasmall{
\textbf{User prompt.} Document: [document]

Criteria: Consider how well each topic captures the general meaning of the document. Consider all types of meaning, including the text's subject matter and the text's affect, emotive content, and style. If Set of Topics A presents a better overall summary of the document compared to Set of Topics B, then "A" should be chosen, and vice versa.

Set of Topics A:

Topic: [topic]

Set of Topics B:

Topic: [topic]
}

\subsection{Examples}

\noindent\textbf{Example 1: Bills with Summarization}

\noindent\extrasmall{
\noindent Document:

\noindent Amends part A of title I of the Elementary and Secondary Education Act of 1965 to alter the percentage of school improvement funds that are reserved for outlying areas and Indian education. Reserves 0.5\% of those funds for outlying areas and 0.75\% for Indian education. (Currently, 1\% of those funds are reserved for outlying areas and Indian education.)

\noindent Set of Topics A: %

\noindent - Educational policies, administration, and support strategies \\
\noindent - Financial management, funding, and policy in public and organizational contexts

\noindent Set of Topics B: %

\noindent - Primary and secondary education systems and funding \\
\noindent - Geographical regions and territories in the United States

\noindent Choice: B

\noindent Reasoning: Set B is more specific to the content of the document, which discusses changes to funding allocations within the Elementary and Secondary Education Act, specifically for outlying areas and Indian education. `Primary and secondary education systems and funding' directly addresses the subject matter, and `Geographical regions and territories in the United States' captures the focus on outlying areas and Indian education. Set A is broader and less directly tied to the specifics of the document.
}

\noindent\textbf{Example 2: PoemSum with TopFeatures}

\noindent\extrasmall{
\noindent Document:

\noindent I saw in Louisiana a live-oak growing, \\
\noindent All alone stood it and the moss hung down from the branches, \\
\noindent Without any companion it grew there uttering joyous leaves of dark green, \\
\noindent And its look, rude, unbending, lusty, made me think of myself, \\
\noindent But I wonder’d how it could utter joyous leaves standing alone there without its friend near, for I knew I could not, \\
\noindent And I broke off a twig with a certain number of leaves upon it, and twined around it a little moss, \\
\noindent And brought it away, and I have placed it in sight in my room, \\
\noindent It is not needed to remind me as of my own dear friends, \\
\noindent (For I believe lately I think of little else than of them,) \\
\noindent Yet it remains to me a curious token, it makes me think of manly love; \\
\noindent For all that, and though the live-oak glistens there in Louisiana solitary in a wide flat space, \\
\noindent Uttering joyous leaves all its life without a friend a lover near, \\
\noindent I know very well I could not.

\noindent Set of Topics A: %

Topic: "tree", "plants", "leaves", "trees", "green", "earth", "boughs", "leaf", "life", "growth"

Topic: "spring", "summer", "autumn", "winter", "fields", "heart", "long", "year", "beauty", "snow"

\noindent Set of Topics B: %

Topic:

\noindent - "literary elements and themes related to poetry"

\noindent - "specific types of flowering plants and their characteristics"

\noindent - "descriptions and characteristics of autumn foliage"

\noindent - "descriptions of agricultural practices and related processes"

\noindent - "references to specific plants and their biochemical properties"

Topic:

\noindent - "emotional connections and feelings of companionship"

\noindent - "themes of longing and escape"

\noindent - "instances of humor or comedic elements in the text"

\noindent - "sentences that imply existential questions or dilemmas regarding trust and character integrity"

\noindent - "expressions related to doubt or uncertainty"

\noindent Choice: B

\noindent Reasoning: Set A focuses on nature, trees, and the cycle of seasons, which are present in the poem, but it misses the deeper emotional and thematic content about solitude, companionship, and longing. Set B, while a bit broad and including some irrelevant elements (like humor and agricultural practices), does include topics about emotional connections, companionship, and existential questions, which are central to the poem's meaning. Despite some extraneous topics, Set B better captures the poem's affective and thematic depth.
}

\normalsize

%% file: sections/appendices/human_evaluation.tex
\section{Human Evaluation Study}
\label{app:human_evaluation}

\begin{figure}[h]
  \centering
  \includegraphics[width=\columnwidth]{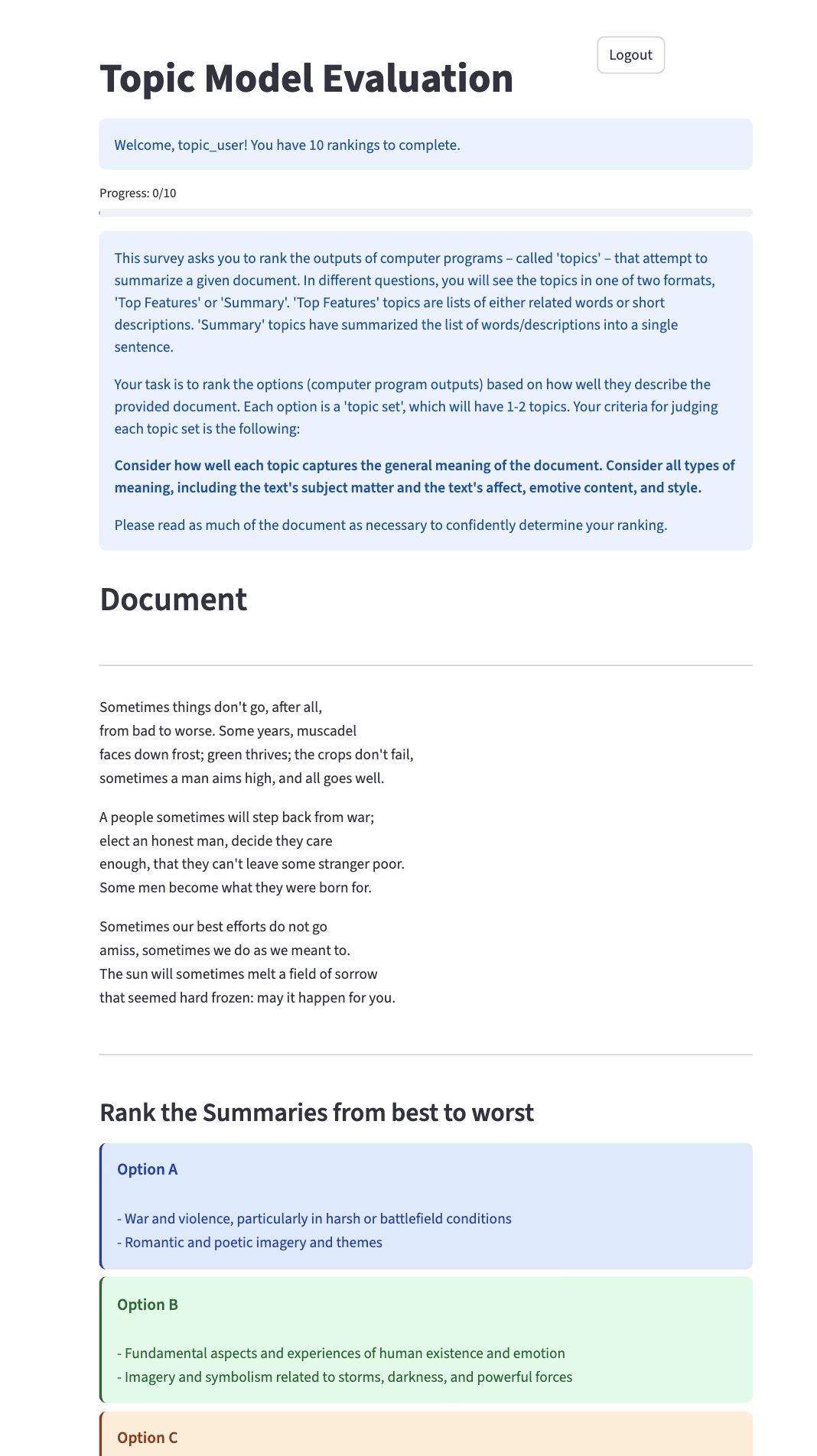}
  \caption{Screenshot of the evaluation interface shown to participants. Each task presents a document alongside topic set options produced by different models. Participants rank the options from best to worst based on how well each topic set captures the document's meaning.}
  \label{fig:human_eval_interface}
\end{figure}

\Cref{fig:human_eval_interface} shows the evaluation interface used in the human topic judge study described in \Cref{sec:empirical_studies:topic_judge}. Below we provide the instructions shown to participants.

\subsection{Instructions Shown to Participants}
\label{app:human_evaluation:instructions}

\extrasmall{
This survey asks you to rank the outputs of computer programs -- called ``topics'' -- that attempt to summarize a given document. In different questions, you will see the topics in one of two formats, ``Top Features'' or ``Summary''. ``Top Features'' topics are lists of either related words or short descriptions. ``Summary'' topics have summarized the list of words/descriptions into a single sentence.

Your task is to rank the options (computer program outputs) based on how well they describe the provided document. Each option is a ``topic set'', which will have 1--2 topics. Your criteria for judging each topic set is the following:

\textbf{Consider how well each topic captures the general meaning of the document. Consider all types of meaning, including the text's subject matter and the text's affect, emotive content, and style.}

Please read as much of the document as necessary to confidently determine your ranking.
}

\normalsize

%% file: sections/appendices/topic_refinement_example.tex
\section{Topic Refinement Example}
\label{app:topic_refinement_example}

\noindent
\begin{minipage}{\linewidth}
\raggedright
\footnotesize{
\begin{itemize}[leftmargin=0pt, itemindent=0pt, label=--, nosep, wide=0pt]
\item \textit{references to Azerbaijani cultural elements, particularly music and instruments}
\item references to protests and related incidents
\item events or actions involving protests and their consequences
\item topics related to the Holocaust and atrocities committed during wartime
\item references to combat and military actions
\item topics related to military actions and warfare
\item references to military involvement or actions related to Russia
\item keywords related to violence and its victims
\item references to violent or aggressive actions
\item instances of violence or conflict
\item \textit{phrases related to archaeological discoveries and remains}
\item mentions of military divisions or actions
\end{itemize}
}

\label{fig:example_topics:mislabeled}
\captionof{figure}{The features identified as spurious in topic refinement for this topic are italicized. The documents associated with this topic discuss the Armenian genocide. The two identified feature descriptions have an incorrect focus (music and instruments) or context (archaeological discoveries).}
\end{minipage}

%% file: sections/appendices/topic-alignment-details.tex
\section{Topic Alignment Procedure for Cross-Correlation Heatmaps}
\label{app:alignment}

To create the cross-correlation heatmaps in \Cref{fig:topic_correlations}, we use greedy alignment:

\begin{enumerate}
\item \textbf{Compute correlations:} For models $m$ and $m'$ with $K$ topics each, compute the $K \times K$ correlation matrix $\mathbf{C}$ where $C_{i,j} = \text{corr}(\theta^m_{:,i}, \theta^{m'}_{:,j})$.

\item \textbf{Greedy matching:} Starting with all topics unmatched, iteratively:
\begin{itemize}
    \item Find the column $j^*$ with highest maximum across all rows: $j^* = \arg\max_{j \in \text{remaining}} \max_i C_{i,j}$
    \item Among remaining rows, find the best match for this column: $i^* = \arg\max_{i \in \text{remaining}} C_{i,j^*}$
    \item Add pair $(i^*, j^*)$ to alignment and remove from consideration
\end{itemize}

\item \textbf{Visualize:} Reorder topics by alignment order to create the heatmap, with strongest column matches appearing first along the diagonal.
\end{enumerate}

High correlations along the diagonal indicate shared concepts between models, while low-correlation regions (dashed boxes) reveal novel topics unique to one model.

%% file: sections/appendices/additional_results.tex
\label{app:additional_results:topical_alignment}

\clearpage
\begin{figure*}[t]
  \centering
  \captionsetup{skip=2pt}
  \renewcommand{\arraystretch}{0.94}
  {\fontsize{6.7pt}{7.3pt}\selectfont
  \begin{tabular}{@{}p{1.1cm}p{12.5cm}@{}}
    \toprule
    \multicolumn{2}{@{}l@{}}{\textbf{Model:} mETM | \textbf{Dataset:} Bills} \\
    \multicolumn{2}{@{}p{13.6cm}@{}}{\textbf{Topic Summary:} "Legal terminology and concepts surrounding property rights, ownership, environmental conservation, and financial transactions in real estate and land management."} \\
    \midrule
    $\lambda=10$ & "A text about the history of a house in New York City, with an emphasis on its architecture and interior design. The author explores how the house has changed over time, from its original construction to its current state [...]" \\
    \midrule
    $\lambda=20$ & "A text about the history of a 10-acre plot in San Bernardino County, California. The property is located near the town of Twentynine Palms and was owned by John W. Smith from 1924 until his death in 1" \\
    \midrule
    $\lambda=30$ & "A text about the 1980s of a parcel in the Town of Silverton, Colorado. The property is located at approximately 35°24'N and 67°W on the west side of Main Street between E.S" \\
    \midrule
    $\lambda=40$ & "A text about 1.02 acres in the town of East Helena, Montana is for sale by sealed bid No. 49-3758 (the "Property") located at approximately 6 miles north and west from the city limits of East" \\
    \midrule
    $\lambda=50$ & "A text about 10 acres in the City of San Bernardino, County of Maricoa.\textbackslash n\textbackslash nThe property is located on a road that was granted by the State of Arizona and United States for the purpose of which is to be used as an irrigation [...]" \\
    \midrule
    \multicolumn{2}{@{}l@{}}{\textbf{Model:} mBERTopic | \textbf{Dataset:} PoemSum} \\
    \multicolumn{2}{@{}p{13.6cm}@{}}{\textbf{Topic Summary:} "Exploration of agriculture, plant varieties, propagation techniques, and their interactions with environmental factors and literary themes."} \\
    \midrule
    $\lambda=10$ & "A text about the history of a house, its inhabitants and their memories.\textbackslash n\textbackslash nThe book is divided into three parts: 1) The story of the house; 2) A series of short stories by different authors from around the world that are inspired [...]" \\
    \midrule
    $\lambda=20$ & "A text about a woman who is in love with her best friend.\textbackslash n\textbackslash nI've been thinking of you all day,\textbackslash nAnd I can't help but wonder if you feel the same way?\textbackslash nYou are my best friend and I know that we will" \\
    \midrule
    $\lambda=30$ & "A text about a girl who was born on the 4th of July, and grew up in the shadow of her mother's love. Her father had died when she was young, but he left behind him a memory that never faded from her heart; for" \\
    \midrule
    $\lambda=40$ & "A text about the beauty of a flower,\textbackslash nThe fragrance that it gives to air;\textbackslash nAnd how its petals fall in showers,\textbackslash nLike stars from heaven's starry hair.\textbackslash n\textbackslash nBut soon they fade and die away,\textbackslash nAs if their hearts were broken" \\
    \midrule
    $\lambda=50$ & "A text about the tree is a beautiful sight to behold.\textbackslash n\textbackslash nThe branches of this tree are full of fruit, and it'ens with blossoms that bloom in the spring. The leaves of this plant are green and red, and they grow on the ground. And" \\
    \midrule
    \multicolumn{2}{@{}l@{}}{\textbf{Unsteered examples} (control)} \\
    \midrule
    \multicolumn{2}{@{}p{13.6cm}@{}}{"A text about a new study on the role of women in science, technology and innovation (STI) was published by UN Women.\textbackslash n\textbackslash nThe report “Women’s participation in STI: A review of evidence” is based on an analysis of 41 studies from"} \\
    \midrule
    \multicolumn{2}{@{}p{13.6cm}@{}}{"A text about the new \emph{Star Wars} movie, which is set to be released in December of this year.\textbackslash n\textbackslash n<blockquote>The film will follow a group of young heroes and villains who are on the run from an evil empire that has taken over their"} \\
    \midrule
    \multicolumn{2}{@{}p{13.6cm}@{}}{"A text about the 1970s, and how it was a time of great change for women.\textbackslash n\textbackslash nThe author talks about her own experience as a young woman in the 70’s, when she had to fight against sexism and discrimination."} \\
    \bottomrule
  \end{tabular}
  }
  \captionof{table}{Representative examples of steered and unsteered text generations for different $\lambda$ values.}
  \label{tab:steered_text_examples}
  
  \vspace{1em}

  \includegraphics[width=0.52\textwidth]{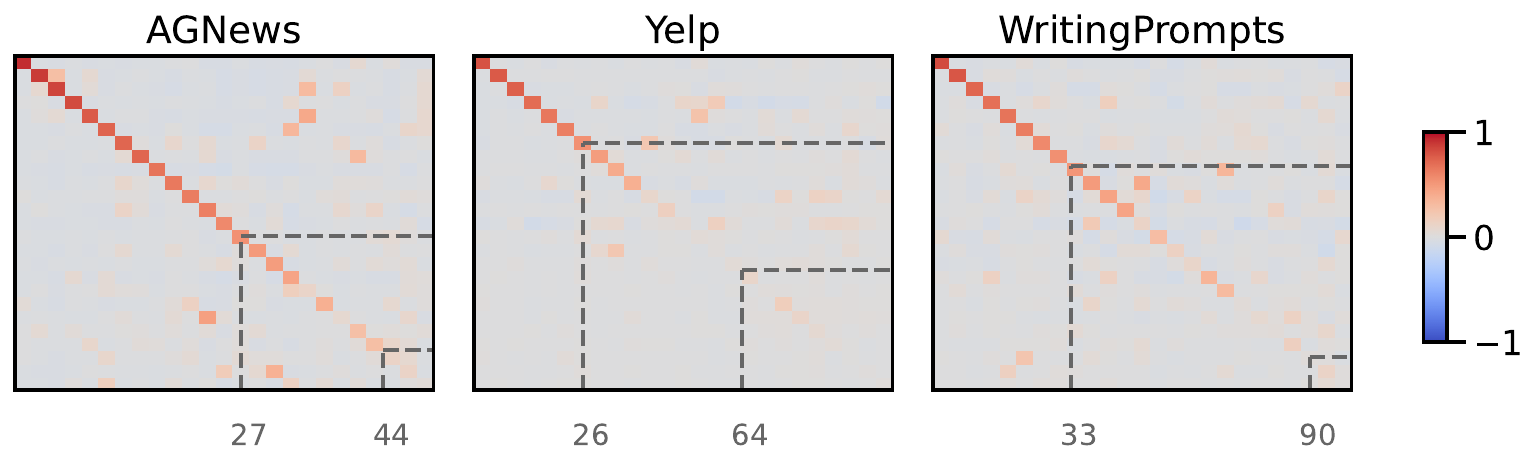}
  \captionof{figure}{Topic cross-correlation heatmaps (mLDA vs.\ LDA) for the three additional datasets. Conventions as in \Cref{fig:topic_correlations}.}
  \label{fig:topic_correlations_new}

  \vspace{1em}

  \makebox[\textwidth][c]{%
    \begin{minipage}[t]{1\textwidth}
      \centering
      \begin{subfigure}[t]{0.98\linewidth}
        \includegraphics[width=\linewidth]{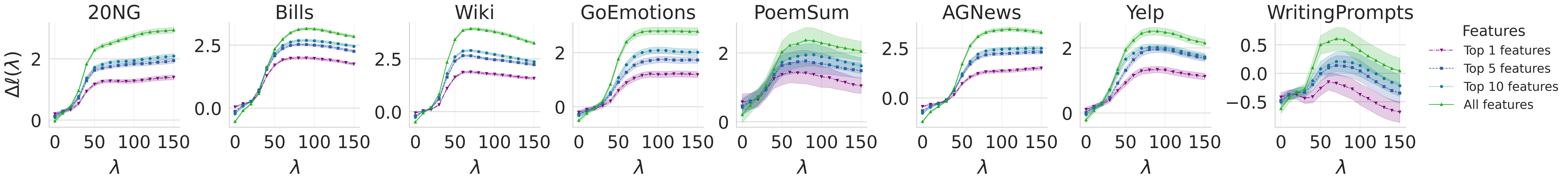}
        \label{fig:etm-log-probs}
      \end{subfigure}

      \vspace{0.2em}

      \begin{subfigure}[t]{0.98\linewidth}
        \includegraphics[width=\linewidth]{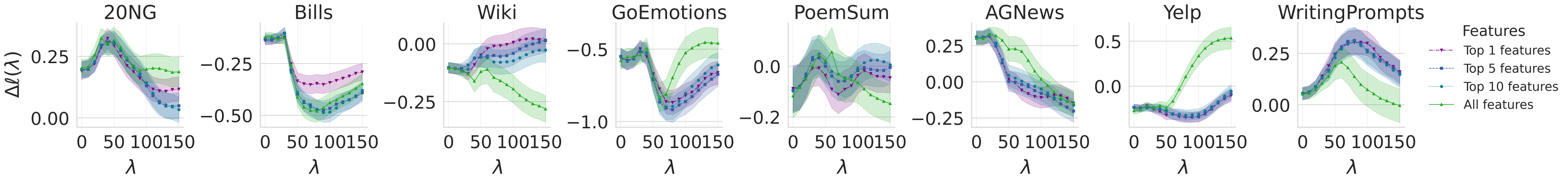}
        \label{fig:lda-log-probs}
      \end{subfigure}
    \end{minipage}%
  }

  \captionof{figure}{Document log-likelihood difference $\Delta\ell(\lambda)$ for mETM (above) and mBERTopic (below). For mETM, $\Delta\ell$ generally increases as $\lambda$ increases, confirming bias toward the target topic. In contrast, mBERTopic does not always show positive $\Delta\ell$ shifts. We think this may be because it is the only model that uses class-based TF-IDF to learn its topic-vocabulary weights, but further investigation is needed. Nevertheless, mBERTopic performs comparably to the other MTMs on metrics measuring the quality of its generated text (\Cref{sec:empirical_studies:steering}), indicating that its steering vectors are still generally effective.}
  \label{fig:combined-log-probs}
\end{figure*}

\clearpage
\begin{figure*}[t]
  \centering
  \captionsetup{skip=2pt}
  \renewcommand{\arraystretch}{0.94}

  \begin{minipage}[t]{0.48\textwidth}
    \centering
    \setlength{\tabcolsep}{2.8pt}%
    \resizebox{\linewidth}{!}{%
      {\fontsize{5.8pt}{6.4pt}\selectfont
      \begin{tabular}{@{}l cccccccc@{}}
        \toprule
        & 20NG & Bills & Wiki & Yelp & AGN. & GoE. & Poem. & WriP. \\
        \midrule
        D-VAE     & .90 & .90 & .86 & .92 & .98 & .96 & .82 & .81 \\
        FASTopic  & \bfseries .99 & \bfseries 1.0 & \bfseries .98 & \bfseries .98 & \bfseries 1.0 & \bfseries 1.0 & \bfseries .99 & \bfseries .96 \\
        \midrule
        LDA       & .61 & .43 & .63 & .75 & .61 & .54 & .47 & .58 \\
        ETM       & .62 & .55 & .71 & .60 & .81 & .79 & .71 & .61 \\
        BERTopic  & .75 & .69 & .65 & .72 & .80 & .78 & .63 & .60 \\
        \midrule
        mLDA      & .71 & .53 & .68 & .53 & .69 & .83 & .64 & .58 \\
        mETM      & .65 & .50 & .61 & .54 & .65 & .81 & .62 & .61 \\
        mBERTopic & .64 & .58 & .60 & .40 & .72 & .85 & .38 & .36 \\
        \bottomrule
      \end{tabular}}
    }
    \captionof{table}{Topic diversity across all datasets, averaged over five runs. Conventions as in \Cref{tab:coherence_td_standard}.}
    \label{tab:coherence_td_short}
  \end{minipage}\hfill
  \begin{minipage}[t]{0.48\textwidth}
    \centering
    \setlength{\tabcolsep}{3.4pt}%
    \resizebox{0.8\linewidth}{!}{%
    {\fontsize{5.8pt}{6.4pt}\selectfont
    \begin{tabular}{@{}l cc cc@{}}
      \toprule
      & \multicolumn{2}{c}{Wiki}
      & \multicolumn{2}{c}{PoemSum} \\
      \cmidrule(lr){2-3} \cmidrule(lr){4-5}
      Model & TF & Sum. & TF & Sum. \\
      \midrule
      D-VAE         & 1383 & 1513 & 1361 & 1376 \\
      FASTopic      & 1387 & 1476 & 1391 & 1424 \\
      \midrule
      LDA           & \cellcolor{losscolbg}1529 & \cellcolor{losscolbg}1574 & \cellcolor{losscolbg}1525 & \cellcolor{losscolbg}1546 \\
      ETM           & \cellcolor{losscolbg}1550 & \cellcolor{losscolbg}1492 & \cellcolor{losscolbg}1471 & \cellcolor{losscolbg}1401 \\
      BERTopic      & \cellcolor{losscolbg}1338 & \cellcolor{losscolbg}1355 & \cellcolor{losscolbg}1487 & \cellcolor{wincolbg}\bfseries1592 \\
      \midrule
      MTM (w/ LDA)  & \cellcolor{wincolbg}\bfseries1657 & \cellcolor{wincolbg}\underline{1574} & \cellcolor{wincolbg}\underline{1619} & \cellcolor{wincolbg}\underline{1550} \\
      MTM (w/ ETM)  & \cellcolor{wincolbg}\underline{1654} & \cellcolor{wincolbg}\bfseries1585 & \cellcolor{wincolbg}\bfseries1627 & \cellcolor{wincolbg}\underline{1561} \\
      MTM (w/ BERTopic) & \cellcolor{wincolbg}1502 & \cellcolor{wincolbg}1430 & \cellcolor{wincolbg}1519 & \cellcolor{losscolbg}\underline{1550} \\
      \bottomrule
    \end{tabular}}
    }
    \captionof{table}{Human topic judge Elo scores on Wiki and PoemSum. Elo conventions as in \Cref{tab:topic_judge_standard}. Human--LLM agreement is visualized in \Cref{fig:human_llm_elo}.}
    \label{tab:human_topic_judge}
    \label{tab:human_llm_agreement}
  \end{minipage}

  \vspace{0.7em}

  \begin{minipage}[t]{0.48\textwidth}
    \centering
    \setlength{\tabcolsep}{3.0pt}%
    \resizebox{\linewidth}{!}{%
      {\fontsize{5.8pt}{6.4pt}\selectfont
      \begin{tabular}{@{}lcccccc@{}}
        \toprule
        & Control & $\lambda=10$ & $\lambda=20$ & $\lambda=30$ & $\lambda=40$ & $\lambda=50$ \\
        \midrule
        mLDA      & 8.2 & 2.8 & 8.4 & 17.9 & 28.0 & 34.5 \\
        mETM      & 10.6 & 3.6 & 8.8 & 16.3 & 27.7 & 33.0 \\
        mBERTopic & 4.8 & 2.4 & 7.4 & 18.2 & 31.5 & 35.8 \\
        \bottomrule
      \end{tabular}}
    }
    \addtocounter{table}{1}
    \captionof{table}{Distribution of TWR wins (\%) by steering magnitude, pooled across all datasets. Each row sums to 100\%. Higher magnitudes account for the majority of steered wins across all models.}
    \label{tab:twr-by-magnitude}
  \end{minipage}\hfill
  \begin{minipage}[t]{0.48\textwidth}
    \centering
    \setlength{\tabcolsep}{2.8pt}%
    \resizebox{\linewidth}{!}{%
    {\fontsize{5.8pt}{6.4pt}\selectfont
    \begin{tabular}{@{}cccccccc@{}}
      \toprule
      20NG & Bills & Wiki & GoEmo. & Poem. & AGN. & Yelp & WriPro. \\
      \midrule
      59.5 & 80.1 & 7.6 & 735.0 & 64.8 & 355.3 & 118.3 & 30.0 \\
      \bottomrule
    \end{tabular}}
    }
    \addtocounter{table}{-2}
    \captionof{table}{Featurization throughput (docs/s) per dataset. We used Gemma~2-9B on an NVIDIA H200 with a batch size of 64 with documents sorted by length. Computations were done on bfloat16 and we only passed the document through the first 10 layers of the LLM as this is the layer used in our experiments.}
    \label{tab:featurization_time}
    \addtocounter{table}{1}
  \end{minipage}

  \vspace{0.7em}

  \begin{minipage}[t]{0.48\textwidth}
    \centering
    \setlength{\tabcolsep}{3.0pt}%
    \resizebox{0.8\linewidth}{!}{%
      {\fontsize{5.8pt}{6.4pt}\selectfont
      \begin{tabular}{@{}lcccc@{}}
        \toprule
        & \multicolumn{2}{c}{Normal} & \multicolumn{2}{c}{Hard} \\
        \cmidrule(lr){2-3}\cmidrule(lr){4-5}
        Model & ${\mathbf{B}}$ & ${\mathbf{\Theta}}$ & ${\mathbf{B}}$ & ${\mathbf{\Theta}}$ \\
        \midrule
        D-VAE           & 0.76 & 0.85 & 0.89 & 0.93 \\
        FASTopic        & 0.70 & 0.79 & 0.83 & 0.86 \\
        \midrule
        LDA             & 0.45 & 0.74 & 0.51 & 0.77 \\
        ETM             & 0.53 & 0.73 & 0.64 & 0.79 \\
        BERTopic        & 0.33 & 0.72 & 0.43 & 0.68 \\
        \midrule
        MTM (w/ LDA)    & \textbf{0.29} & \textbf{0.48} & \textbf{0.32} & \textbf{0.49} \\
        MTM (w/ ETM)    & 0.35 & 0.55 & 0.40 & 0.54 \\
        MTM (w/ BERTopic) & 0.30 & 0.74 & 0.35 & 0.69 \\
        \bottomrule
      \end{tabular}}
    }
    \captionof{table}{Averaged model stability for normal (20NG, Bills, Wiki, Yelp, AGN) and hard (GoEmo., PoemSum, WriPro.) datasets. Each model is trained 5 times with different random seeds. Distance between topic pairs is defined as the Rank-Biased Overlap distance ($p=0.9$) of the top 500 features ($\mathbf{B}$) or documents ($\mathbf{\Theta}$); topics are then matched with the Hungarian algorithm \citep{hoyle-etal-2022-neural}. MTMs are overall more stable than their word-based counterparts.}
    \label{tab:stability}
  \end{minipage}\hfill
  \begin{minipage}[t]{0.48\textwidth}
    \centering
    \vspace*{-5.8em}
    \setlength{\tabcolsep}{2.5pt}%
    \resizebox{\linewidth}{!}{%
      {\fontsize{5.6pt}{6.2pt}\selectfont
      \begin{tabular}{@{}lcccccccc@{}}
        \toprule
        & 20NG & Bills & Wiki & Yelp & AGN & GoEmo & Poem & WP \\
        \midrule
        TWR (\%)                     & 93.9 & 99.2 & 96.5 & 98.1 & 99.2 & 93.7 & 97.0 & 85.3 \\
        \midrule
        $\text{PPL}_\text{control}$  & 6.25 & 6.21 & 6.22 & 6.21 & 6.19 & 6.16 & 6.24 & 6.24 \\
        $\text{PPL}_{\lambda=10}$    & 6.41 & 6.34 & 6.53 & 6.50 & 6.13 & 6.25 & 6.23 & 6.22 \\
        $\text{PPL}_{\lambda=20}$    & 7.00 & 6.86 & 6.91 & 7.39 & 6.73 & 6.76 & 7.10 & 7.00 \\
        $\text{PPL}_{\lambda=30}$    & 8.10 & 8.02 & 7.76 & 8.96 & 7.42 & 7.81 & 8.93 & 8.29 \\
        $\text{PPL}_{\lambda=40}$    & 10.02 & 9.75 & 9.24 & 11.11 & 8.47 & 10.46 & 11.46 & 11.25 \\
        $\text{PPL}_{\lambda=50}$    & 15.01 & 12.73 & 12.43 & 15.37 & 11.30 & 23.52 & 16.48 & 25.34 \\
        \bottomrule
      \end{tabular}}
    }
    \captionof{table}{Steering metrics for mBERTopic across all 8 datasets. Conventions as in \Cref{tab:winrates}.}
    \label{tab:mbertopic-twr-ppl-results}
    \vspace*{0.9em}
  \end{minipage}

  \vspace{0.8em}

  \begin{minipage}[t]{0.48\textwidth}
    \centering
    \setlength{\tabcolsep}{2.5pt}%
    \resizebox{\linewidth}{!}{%
      {\fontsize{5.5pt}{6.1pt}\selectfont
      \begin{tabular}{@{}lcccccccc@{}}
        \toprule
        & \multicolumn{2}{c}{20NG} & \multicolumn{2}{c}{Bills} & \multicolumn{2}{c}{Wiki} & \multicolumn{2}{c}{GoEmotions} \\
        \cmidrule(lr){2-3}\cmidrule(lr){4-5}\cmidrule(lr){6-7}\cmidrule(lr){8-9}
        Model & $P_1$ & NMI & $P_1$ & NMI & $P_1$ & NMI & $P_1$ & NMI \\
        \midrule
        D-VAE           & 0.27 & 0.33 & 0.33 & 0.38 & \underline{0.45} & \underline{0.71} & 0.28 & 0.05 \\
        FASTopic        & 0.58 & 0.47 & 0.44 & 0.48 & 0.41 & 0.65 & 0.29 & 0.06 \\
        \midrule
        LDA             & 0.54 & 0.44 & 0.50 & 0.51 & 0.44 & \bfseries 0.72 & 0.28 & 0.05 \\
        ETM             & 0.54 & 0.45 & 0.46 & 0.50 & 0.36 & 0.69 & 0.30 & 0.05 \\
        BERTopic        & \bfseries 0.65 & \bfseries 0.59 & 0.49 & \bfseries 0.56 & 0.42 & 0.69 & 0.27 & 0.04 \\
        \midrule
        MTM (w/ LDA)    & 0.49 & 0.42 & \bfseries 0.52 & 0.53 & 0.42 & 0.71 & \bfseries 0.31 & \bfseries 0.08 \\
        MTM (w/ ETM)    & 0.53 & 0.44 & 0.51 & 0.53 & \bfseries 0.45 & 0.71 & \underline{0.31} & 0.08 \\
        MTM (w/ BERTopic) & 0.56 & 0.51 & 0.47 & 0.53 & 0.37 & 0.68 & 0.28 & 0.07 \\
        \bottomrule
      \end{tabular}}
    }
    \captionof{table}{Alignment metrics Purity ($P_1$) and NMI for labeled datasets, averaged over five runs. Conventions as in Table~2. Purity quantifies single-category clusters; NMI measures topic-label mutual information \citep{hoyle-etal-2022-neural}. MTMs outperform word-based counterparts on GoEmotions, consisting of short texts labeled with one of 28 emotions.}
    \label{tab:alignment}
  \end{minipage}\hfill
  \begin{minipage}[t]{0.48\textwidth}
    \centering
    \vspace*{-15.2em}
    \setlength{\tabcolsep}{2.5pt}%
    \resizebox{\linewidth}{!}{%
      {\fontsize{5.6pt}{6.2pt}\selectfont
      \begin{tabular}{@{}lcccccccc@{}}
        \toprule
        & 20NG & Bills & Wiki & Yelp & AGN & GoEmo & Poem & WP \\
        \midrule
        TWR (\%)                     & 88.1 & 99.2 & 93.1 & 88.7 & 93.0 & 76.0 & 85.8 & 84.9 \\
        \midrule
        $\text{PPL}_\text{control}$  & 6.20 & 6.20 & 6.25 & 6.21 & 6.27 & 6.20 & 6.22 & 6.21 \\
        $\text{PPL}_{\lambda=10}$    & 6.41 & 6.17 & 6.41 & 6.44 & 6.18 & 6.27 & 6.31 & 6.30 \\
        $\text{PPL}_{\lambda=20}$    & 6.83 & 6.85 & 7.03 & 7.19 & 6.66 & 6.75 & 7.05 & 6.84 \\
        $\text{PPL}_{\lambda=30}$    & 7.93 & 8.02 & 7.74 & 8.74 & 7.70 & 7.80 & 8.66 & 8.10 \\
        $\text{PPL}_{\lambda=40}$    & 10.04 & 9.66 & 9.26 & 11.11 & 8.72 & 10.83 & 11.51 & 10.81 \\
        $\text{PPL}_{\lambda=50}$    & 14.40 & 12.87 & 12.88 & 16.93 & 11.83 & 21.06 & 17.33 & 17.02 \\
        \bottomrule
      \end{tabular}}
    }
    \captionof{table}{Steering metrics for mETM across all 8 datasets. Conventions as in \Cref{tab:winrates}.}
    \label{tab:all-twr-ppl-results}

    \vspace{1.5em}

    \begin{minipage}[c]{0.08\linewidth}
      \centering
      \rotatebox[origin=c]{90}{\tiny $\Delta \ell(\lambda)$}
    \end{minipage}\hspace{0.01\linewidth}%
    \begin{minipage}[c]{0.89\linewidth}
      \includegraphics[width=\linewidth,trim=1435 0 0 0,clip]{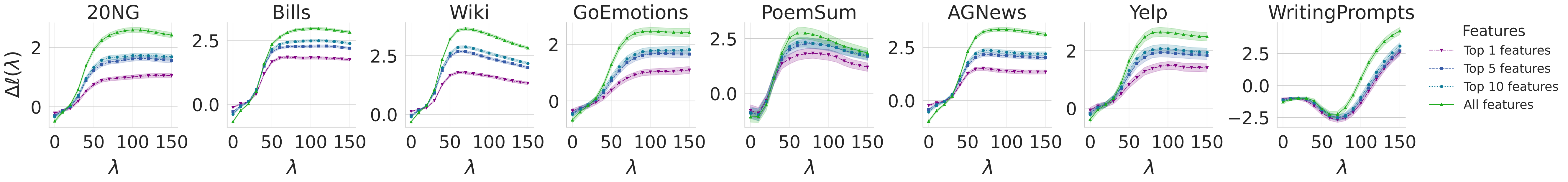}
    \end{minipage}
    \captionof{figure}{Document log-likelihood difference $\Delta\ell(\lambda)$ for mLDA on AGNews, Yelp, and WritingPrompts. As in \Cref{fig:log-prob-graphs}, $\Delta\ell$ generally becomes more positive as steering strengthens, with the full steering vector producing the largest shift.}
    \label{fig:mlda-log-probs-right-third}
  \end{minipage}
\end{figure*}

%% file: sections/appendices/topic_examples.tex
\FloatBarrier
\clearpage
\begin{figure*}[!t]
  \centering
  \captionsetup{skip=2pt}
  \captionsetup[subfigure]{skip=2pt}
  \lstset{aboveskip=0.5ex,belowskip=0.5ex}
  \scriptsize
  \setlength{\tabcolsep}{4pt}

  \begin{minipage}[t]{0.48\textwidth}
    \begin{subfigure}[t]{\linewidth}
      \begin{lstlisting}[basicstyle=\ttfamily\tiny]
species, shark, cap, stem, sharks, brown, fungus, shaped, smooth, fruit, spores, fins, bodies, surface, genus, typically, thick, distribution, habitat, fin\end{lstlisting}
      \begin{lstlisting}[basicstyle=\ttfamily\scriptsize]
- scientific classifications and descriptions of plant species
- terms related to flora and plant characteristics
- details about plant characteristics and descriptions
- terms related to cryptic species and their ecological studies
- specific colors and descriptions related to plants and their characteristics
- terms related to biological and anatomical features\end{lstlisting}
      \caption{[Wiki] mETM uncovers high-level botanical taxonomy concepts, robust to varied word choice.}
    \end{subfigure}

    \vspace{0.5em}

\begin{subfigure}[t]{\linewidth}
\begin{lstlisting}[basicstyle=\ttfamily\scriptsize]
people, law, government, right, rights, laws, like, case, think, state, public, crime, use, time, person\end{lstlisting}
     \begin{lstlisting}[basicstyle=\ttfamily\scriptsize]
- negative sentiments directed towards authority and governance
- references to freedom of speech and expression
- references to extremist ideologies and discriminatory language
- names of political figures and references to political actions
- themes related to political criticism and party dynamics
- phrases related to political accountability and ethics\end{lstlisting}
    \caption{[20NG] mLDA captures both critical tone and content in political discourse, whereas LDA captures only the broader political theme.}
    \end{subfigure}

    \vspace{0.5em}

    \begin{subfigure}[t]{\linewidth}
      \begin{lstlisting}[basicstyle=\ttfamily\scriptsize]
- expressions of disbelief or surprise
- expressions of unexpectedness or surprise in various contexts
- words expressing strangeness or oddity
- expressions of emotions and connections between people
- expressions of doubt, reflection, and introspection
- expressions of surprise or shock in personal experiences\end{lstlisting}
    \caption{[GoEmotions] mLDA identifies complex emotional states (surprise, disbelief) as a coherent topic.}
    \end{subfigure}

    \vspace{0.5em}

    \begin{subfigure}[t]{\linewidth}
      \begin{lstlisting}[basicstyle=\ttfamily\scriptsize]
- references to relationship dynamics and communication issues
- references to workplace disputes and grievances
- familial relationships and conflicts
- references to psychological distress and coping mechanisms
- phrases and terms related to emotional support and effective communication in caregiving
- emotions related to anger and frustration\end{lstlisting}
      \caption{[20NG] mBERTopic reveals nuanced interpersonal conflict and coping themes.}
    \end{subfigure}

    \vspace{0.5em}

    \begin{subfigure}[t]{\linewidth}
      \begin{lstlisting}[basicstyle=\ttfamily\scriptsize]
- expressions related to rudeness and offensive behavior
- expressions of anger and strong negative emotions
- terms associated with negative character traits or behavior
- expressions of criticism and critical comments
- themes related to feelings of embarrassment and shame
- references to feedback, accountability, and constructive criticism\end{lstlisting}
      \caption{[GoEmotions] mLDA captures sentiment and tone in a unified topic, identifying negative sentiment and social disapproval themes.}
    \end{subfigure}

  \end{minipage}%
  \hfill
  \begin{minipage}[t]{0.48\textwidth}
    \begin{subfigure}[t]{\linewidth}
      \begin{lstlisting}[basicstyle=\ttfamily\scriptsize]
government, political, minister, party, military, prime, president, national, economic, leader, country, opposition, policy, parliament, foreign, secretary, leadership, independence, general, leaders\end{lstlisting}
      \begin{lstlisting}[basicstyle=\ttfamily\scriptsize]
- references to historical or political movements and conflicts
- references to political dynamics, power struggles, and social discrimination
- references to specific groups or organizations
- references to U.S.-backed interventions and coups in foreign nations
- references to political factions or militia groups involved in conflicts
- references to historical events involving the Soviet Union\end{lstlisting}
      \caption{[Wiki] mETM provides rich contextual descriptions of historical conflicts versus ETM's generic political terms.}
    \end{subfigure}

    \vspace{0.5em}

    \begin{subfigure}[t]{\linewidth}
      \begin{lstlisting}[basicstyle=\ttfamily\scriptsize]
- references to characters or individuals from Greek mythology
- nouns associated with historical events and figures
- instances of significant literary expressions or metaphors
- mythological figures and their related narratives
- religious or spiritual references and concepts
- references to mythological gods and their interactions\end{lstlisting}
      \caption{[PoemSum] mETM discovers novel Greek mythology theme (max correlation 0.12 with any ETM topic).}
    \end{subfigure}

    \vspace{0.5em}

    \begin{subfigure}[t]{\linewidth}
      \begin{lstlisting}[basicstyle=\ttfamily\scriptsize]
beauty, woman, greece, maid, thy, praise, eyes, white, lovely, thee, world, skill, face, darkened, love, wan, fleet, deeds, lip, art\end{lstlisting}
      \begin{lstlisting}[basicstyle=\ttfamily\scriptsize]
- concepts related to existential purpose and the divine
- themes related to existential questions and the nature of God
- instances of significant literary expressions or metaphors
- literary elements and themes related to poetry
- concepts related to spirituality and eternal life
- descriptive phrases related to physical characteristics or appearances\end{lstlisting}
      \caption{[PoemSum] mBERTopic captures existential and religious motifs beyond surface-level poetic vocabulary.}
    \end{subfigure} 
    
 \vspace{0.75em}
 
    \begin{subfigure}[t]{\linewidth}
      \begin{lstlisting}[basicstyle=\ttfamily\scriptsize]
- literary elements and themes related to poetry
- information about an artist's biography and professional background
- emotional connections and feelings of companionship
- references to poetry and poets
- instances of humor or comedic elements in the text
- sentences that imply existential questions or dilemmas regarding trust and character integrity
- keywords or phrases related to judicial or legal processes
- mentions of struggles and hardships faced by individuals
\end{lstlisting}
      \caption{[PoemSum] mLDA demonstrates a potential MTM failure mode: while the features correctly represent a poetic theme, they're too broad to be useful, and spurious features irrelevant to the topic's top documents persist despite post-training refinement (\Cref{app:posttraining_refinement}).}
    \end{subfigure}
  \end{minipage}
  \vspace{0.75em}
  \caption{
    Illustrative examples of topic sets from MTMs (mLDA, mETM, mBERTopic) across diverse datasets. Each mechanistic topic is represented by its top 6--8 SAE features. When appropriate, mechanistic topics are contrasted with their most similar word-based topic (top 20 words), selected based on correlation of document-topic distributions (see \Cref{sec:empirical_studies:qualitative}). The word-based topic is shown above the corresponding mechanistic topic. 
  }
  \label{fig:new-topics}
\end{figure*}